\documentclass[11pt]{article}

\usepackage[final]{acl}

\usepackage{times}
\usepackage{latexsym}

\usepackage[T1]{fontenc}

\usepackage[utf8]{inputenc}

\usepackage{microtype}

\usepackage{inconsolata}

\usepackage{graphicx}

\usepackage{booktabs}
\usepackage{adjustbox}
\usepackage{caption}
\usepackage{multirow}
\usepackage{xcolor}
\usepackage{siunitx}

\usepackage{graphicx}
\usepackage{subcaption}

\usepackage{textcomp}
\usepackage{tabularx}

\usepackage[table]{xcolor}
\usepackage{multirow}
\usepackage{amssymb} 
\usepackage{makecell}
\usepackage{enumitem}

\usepackage{array}
\usepackage{arydshln}

\usepackage{placeins}
\usepackage{threeparttable}
\usepackage[dvipsnames]{xcolor}
\definecolor{myred}{HTML}{C00000}
\definecolor{myblue}{HTML}{002060}
\newcommand{\dashsep}{\noalign{\vskip 0.15em}}

\usepackage{xcolor}
\usepackage{pifont} 
\newcommand{\cmark}{\ding{51}} 
\newcommand{\xmark}{\ding{55}} 
\newcommand{\good}{\textcolor{green!60!black}{\cmark}}
\newcommand{\bad}{\textcolor{red!70!black}{\xmark}}

\usepackage{booktabs}
\usepackage{siunitx}
\usepackage{multirow}
\usepackage{adjustbox}
\usepackage[table]{xcolor}
\usepackage{colortbl}

\usepackage{balance}

\definecolor{rowblue}{RGB}{235,245,255}

\definecolor{subgray}{gray}{0.96}       

\definecolor{cot-green}{RGB}{126,172,85}
\definecolor{cot-red}{RGB}{234,51,35}
\definecolor{cot-blue}{RGB}{79,113,190}
\newcolumntype{M}[1]{>{\centering\arraybackslash}m{#1}}
\title{Cross-Modal Coreference Alignment: Enabling Reliable Information Transfer in Omni-LLMs}

\author{
Hongcheng Liu\textsuperscript{*}, 
Yuhao Wang\textsuperscript{*}, 
Zhe Chen, 
Pingjie Wang, 
Zhiyuan Zhu \\
{\bf Yixuan Hou, 
Yanfeng Wang, 
Yu Wang\textsuperscript{\dag}} \\
Shanghai Jiao Tong University \\
\texttt{\{hongcheng\_liu, colane, yuwangsjtu\}@sjtu.edu.cn}\\
\small{\textsuperscript{*} Equal contribution. \quad
\textsuperscript{\dag} Corresponding author.}
}

\begin{document}
\maketitle
\begin{abstract}
Omni Large Language Models (Omni-LLMs) have demonstrated impressive capabilities in holistic multi-modal perception, yet they consistently falter in complex scenarios requiring synergistic omni-modal reasoning. Beyond understanding global multimodal context, effective reasoning also hinges on fine-grained cross-modal alignment, especially identifying shared referents across modalities, yet this aspect has been largely overlooked. To bridge this gap, we formalize the challenge as a cross-modal coreference problem, where a model must localize a referent in a source modality and re-identify it in a target modality. Building on this paradigm, we introduce \textsc{CrossOmni}, a dataset comprising nine tasks equipped with human-designed reasoning rationales to evaluate and enhance this capability. Experiments on 13 Omni-LLMs reveal systematic weaknesses in cross-modal coreference, which we attribute to the absence of coreference-aware thinking patterns. To address this, we enhance cross-modal alignment via two strategies: a training-free In-Context Learning method and a training-based SFT+GRPO framework designed to induce such thinking patterns. Both approaches yield substantial performance gains and generalize effectively to collaborative reasoning tasks. Overall, our findings highlight cross-modal coreference as a crucial missing piece for advancing robust omni-modal reasoning.
\end{abstract}
\section{Introduction}

Omni Large Language Models (Omni-LLMs) have demonstrated impressive capabilities in holistic multi-modal perception, yet they consistently falter in complex scenarios requiring synergistic omni-modal reasoning~\cite{liu2025ola,yao2024minicpm}. Most prior work attempts to improve it by aggregating information from multiple modalities, such as expanding data coverage~\cite{ye2025omnivinci,ai2025ming} or refining training objectives~\cite{yang2025humanomniv2,zhong2025omni}. However, effective reasoning requires not only modeling global multimodal context but also performing fine-grained cross-modal alignment~\cite{li2025alignmamba}. Models must bind the same referent across modalities and align its evolving state to compose evidence into coherent reasoning. While aggregation-oriented methods can enhance overall perception, they often underemphasize such alignment and fail to reliably link corresponding content across modalities. Consequently, current Omni-LLMs struggle to maintain referential consistency~\cite{alonso2025vision}, which severely constrains their reasoning performance.

\begin{figure}[t]
    \centering
    \includegraphics[width=0.98\linewidth]{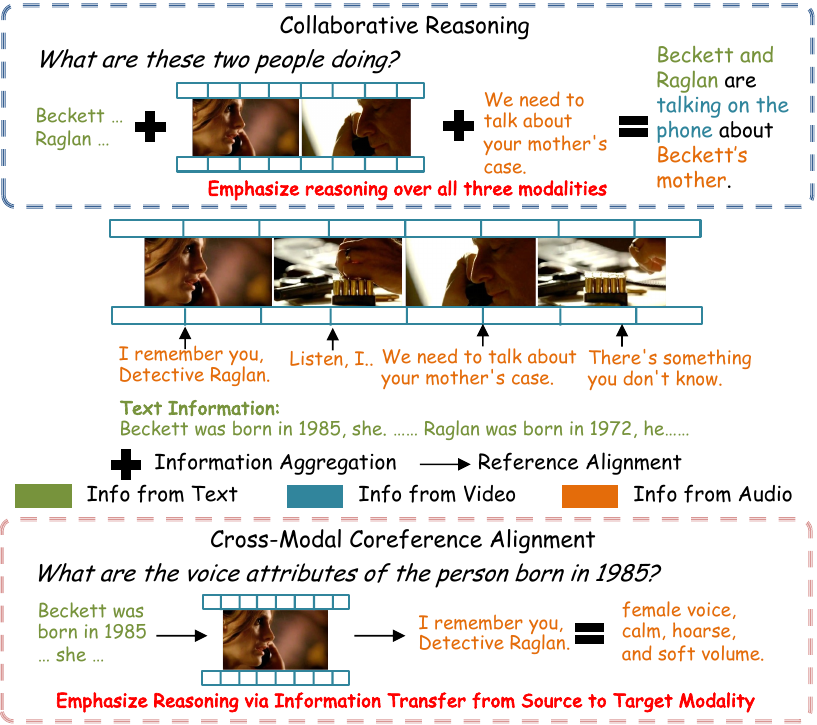}
    \caption{The comparison between collaborative reasoning and cross-modal coreference alignment. The former focuses on coarse aggregation, whereas the latter targets fine-grained alignment of corresponding referents.}
    \label{fig:figure1}
    \vspace{-1.2em}
\end{figure}

\begin{table*}[t]
\centering
\renewcommand{\baselinestretch}{1.0}
\begin{adjustbox}{max width=0.93\textwidth}
\begin{tabular}{lccccccc}
\toprule
\bf Datasets & \bf \# Videos  & \bf \# QA Pairs & \bf Modality & \bf w. Cross-Modality & \bf Cross-Modal Coreference & \bf w. Training Data \\
\midrule
Pano-AVQA~\citeyearpar{yun2021pano}  & 5,400 & 51,700 & A, V & \bad & - & \good  \\
AVQA~\citeyearpar{yang2022avqa} & 57,015 & 57,335 & A ,V & \bad & - & \good \\
Music-AVQA~\citeyearpar{music-avqa} & 9,288 & 45,867 & A, V & \bad & - & \good \\
OmniBench~\citeyearpar{yizhiomnibench} & - & 4,555 & A, I & \bad & - & \bad \\
WorldSense~\citeyearpar{hong2025worldsense} & 1,662 & 3,172 & A, V & \bad & - & \bad \\
DailyOmni~\citeyearpar{zhou2025dailyomni} & 684 & 1,197 & A, V & \bad & - & \bad \\
AVUT~\citeyearpar{yang2025avut} & 2,662  & 11,609 & A, V & \good & A$\to$V & \bad \\
\midrule
\textsc{\textbf{CrossOmni}}& 4,147  & 39,726 & A, V, T & \good & A$\leftrightarrow$V, A$\leftrightarrow$T, V$\leftrightarrow$T & \good \\
\bottomrule
\end{tabular}
\end{adjustbox}
\caption{Comparison of \textsc{CrossOmni} with representative omni-modal datasets. Cross-modality denotes questions that require information transfer across modalities. A/I/V/T denote audio, image, video, and text.}
\label{tab:dataset_comparasion}
\vspace{-1.5em}
\end{table*}

To bridge this gap, we formalize the challenge as a cross-modal coreference problem. Moving beyond the collaborative reasoning focused on coarse aggregation in prior work, our approach targets a critical intermediary step: precisely locating a referent in a source modality and re-identifying it within a target modality (Figure~\ref{fig:figure1}). We introduce \textsc{CrossOmni}, a comprehensive dataset designed to evaluate and elicit this capability. \textsc{CrossOmni} encompasses three single-modality and six cross-modality coreference tasks spanning text, audio, and video. Crucially, it provides human-annotated reasoning rationales that supervise step-wise alignment, facilitating both granular evaluation and effective model tuning.

Benchmarking 13 leading Omni-LLMs reveals a persistent performance gap between single- and cross-modal coreference, highlighting a systemic deficiency. Through Chain-of-Thought (CoT) prompting, we observe that models often lack the necessary ``thinking patterns'' to transfer information reliably between modalities. Motivated by this, we explore two strategies to induce coreference-aware thinking: a training-free In-Context Learning approach and a training-based method combining SFT with Group Relative Policy Optimization (GRPO). Both methods yield significant gains across all coreference tasks. Furthermore, our ablation studies demonstrate that these improvements generalize to broader collaborative reasoning benchmarks, suggesting that cross-modal coreference is a foundational ``missing piece'' in the evolution of Omni-LLM intelligence.

The main contributions can be summarized as:
\begin{itemize}[itemsep=0pt, topsep=0pt, parsep=0pt]
    \item \textbf{Cross-Modal Coreference Alignment.}
    We formulate cross-modal coreference alignment, where the model must identify information in a target modality by first locating it in a source modality. This paradigm shifts the focus for improving Omni-LLMs from coarse modality aggregation to fine-grained cross-modal coreference alignment.
    \item \textbf{Rationale-Augmented Multi-Task Dataset.} This dataset comprises 9 diverse task types enriched with human-designed rationales, specifically developed to facilitate the training and evaluation of Omni-LLMs in complex cross-modal coreference alignment.
    \item \textbf{Coreference-aware Thinking Patterns.} We introduce both novel training-free and training-based methods to elicit structured thinking patterns. We show that these patterns significantly boost performance on both coreference-specific tasks and general collaborative reasoning benchmarks.
\end{itemize}
\section{\textsc{CrossOmni} Dataset}
\begin{figure*}[t]
    \centering
    \includegraphics[width=0.98\linewidth]{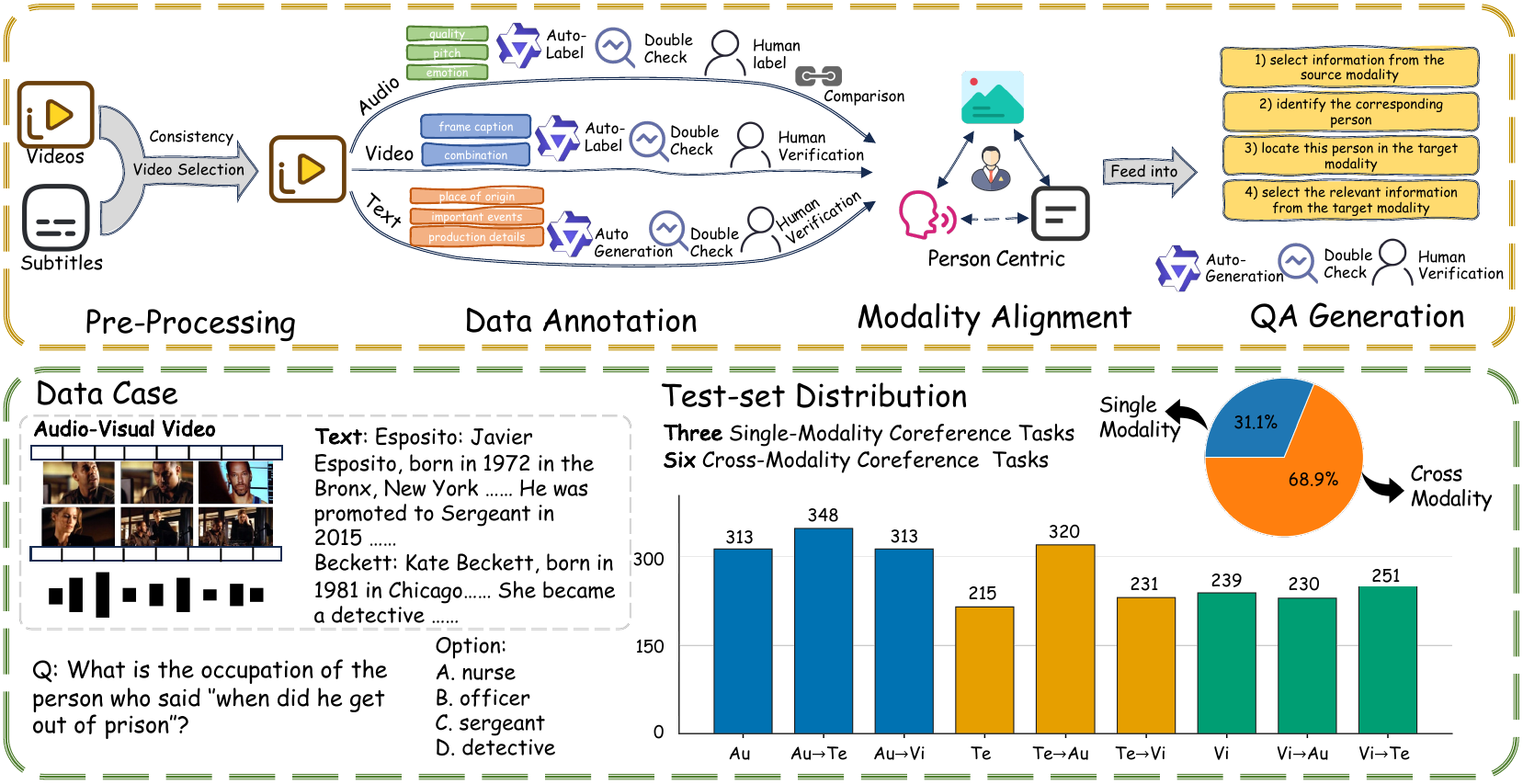}
    \caption{Overview of the \textsc{CrossOmni} dataset. The top section illustrates the annotation pipeline, including pre-process, data annotation, modality alignment, and QA generation. The bottom section presents example instances and the distribution of the test set.}
    \label{fig:pipeline}
    \vspace{-0.8em}
\end{figure*}

\subsection{Dataset Overview}
\subsubsection{Data Principles}
The \textsc{CrossOmni} dataset is designed to evaluate and support training for cross-modal coreference alignment over three modalities in Omni-LLMs. Unlike existing omni benchmarks that simply fuse information from different modalities and aggregate various cues to answer a question, \textsc{CrossOmni} emphasizes cross-modality coreference among audio, visual, and textual data, as illustrated in Figure~\ref{fig:pipeline}. 
Specifically, \textsc{CrossOmni} pairs single-modality coreference tasks with cross-modality coreference tasks, allowing unimodal performance to serve as a baseline for directly assessing and highlighting cross-modal evidence transfer within a unified framework.

Concretely, we construct a diverse set of videos with comprehensive annotations, providing training sets and test splits for cross-modality alignment, consisting of 4{,}147 videos and 39{,}726 question–answer pairs (Table~\ref{tab:dataset-split}). All videos are sampled from TVQA\footnote{\url{https://nlp.cs.unc.edu/data/jielei/tvqa/tvqa_public_html/index.html}} and come from six long-running TV shows with multi-person conversations and complex visual scenes, which pose significant challenges for cross-modality coreference alignment. For each modality, we perform detailed annotation and automatic verification, followed by human checking to ensure data quality. We also provide explicit step-by-step rationales for each instance, offering effective thinking patterns that enable more precise analysis and training of cross-modal coreference alignment.

\begin{table}[t]
  \centering
  \small
  \begin{threeparttable}
  \setlength{\tabcolsep}{10pt}
  \begin{tabular}{lccc}
    \toprule
    \textbf{Type} & \textbf{\# Train} & \textbf{\# Test} & \textbf{\# Total} \\
    \midrule
    Videos   & 2,725 & 1,422 & 4,147 \\
    QA pairs & 37,266 & 2,460 & 39,726 \\
    \bottomrule
  \end{tabular}
   \caption{Statistics of the \textsc{CrossOmni} dataset.}
  \label{tab:dataset-split}
  \end{threeparttable}
  \vspace{-1.5em}
\end{table}

\subsubsection{Task Definition}
We model each instance as a source-to-target coreference alignment problem.
Given a question $q$, a source modality $m_s \in \{\text{Au},\text{Vi},\text{Te}\}$, and a target modality $m_t \in \{\text{Au},\text{Vi},\text{Te}\}$, the model is required to:
(1) identify the referent in the source modality that is needed to answer $q$, and
(2) retrieve the aligned information in the target modality to produce the final answer $a$.
Single-modality coreference corresponds to $m_s = m_t$, while cross-modality coreference corresponds to $m_s \neq m_t$.

\subsubsection{Task Classification}
Under task definition, we define nine task types: three single-modality coreference tasks and six cross-modality coreference tasks over audio, video, and text. Table~\ref{tab:tasks_cases} presents representative examples for each type. The information available in each modality is summarized below, and more details are provided in Appendix~\ref{app:task_detail}.
\begin{table*}[t]
  \centering
  \small
  \setlength{\dashlinedash}{0.5pt}
  \setlength{\dashlinegap}{2pt}
  \setlength{\tabcolsep}{6pt}
  \renewcommand{\arraystretch}{1.18}
  \begin{tabularx}{0.98\textwidth}{@{}>{\raggedright\arraybackslash}p{0.21\textwidth} X@{}}
    \toprule
    \multicolumn{2}{c}{\textbf{Single-modality Coreference Tasks}} \\
    \midrule

    \textbf{Audio-only (Au)} &
      \makecell[l]{What is the emotion in the sentence ``why would she do that''?\\
      A. calm \quad B. excited \quad C. angry \quad D. sad} \\
    \dashsep
    \hdashline
    \dashsep

    \textbf{Vision-only (Vi)} &
      \makecell[l]{Where does the third woman enter from?\\
      A. front side \quad B. back side \quad C. left side \quad D. right side} \\
    \dashsep
    \hdashline
    \dashsep

    \textbf{Text-only (Te)} &
      \makecell[l]{What was beckett's first screenplay?\\
      A. the weight of silence \quad B. silent echoes \quad
      C. the weight of words \quad D. echoes of the past} \\

    \midrule
    \multicolumn{2}{c}{\textbf{Cross-modality Coreference Tasks}} \\
    \midrule

    \textbf{Audio$\rightarrow$Vision (Au$\rightarrow$Vi)} &
      \makecell[l]{What facial expression does the person who spoke with aversion emotion exhibit?\\
      A. disoriented look \quad B. calm expression \quad \\
      C. serious or concerned expression \quad D. attentive look} \\
    \dashsep
    \hdashline
    \dashsep

    \textbf{Audio$\rightarrow$Text (Au$\rightarrow$Te)} &
      \makecell[l]{What is the occupation of the person who said ``when did he get out of prison''?\\
      A. nurse \quad B. officer \quad C. sergeant \quad D. detective} \\
    \dashsep
    \hdashline
    \dashsep

    \textbf{Vision$\rightarrow$Audio (Vi$\rightarrow$Au)} &
      \makecell[l]{What does the person's emotion in voice who has wide eyes and open mouth in the video?\\
      A. sadness \quad B. excitement \quad C. happiness \quad D. fear} \\
    \dashsep
    \hdashline
    \dashsep

    \textbf{Vision$\rightarrow$Text (Vi$\rightarrow$Te)} &
      \makecell[l]{What is the birthdate of the person wearing brown pants and a black t-shirt?\\
      A. april 5, 1985 \quad B. january 1, 1985 \quad
      C. march 12, 1985 \quad D. march 12, 1984} \\
    \dashsep
    \hdashline
    \dashsep

    \textbf{Text$\rightarrow$Audio (Te$\rightarrow$Au)} &
      \makecell[l]{What does the person say who joined the princeton-plainsboro teaching hospital?\\
      A. I'd be too nervous. Couldn't be objective. \quad
      B. I need to know everything about you.\\
      C. here's what happens when doctors care too much. \quad
      D. She's mine. Stay away.} \\
    \dashsep
    \hdashline
    \dashsep

    \textbf{Text$\rightarrow$Vision (Te$\rightarrow$Vi)} &
      \makecell[l]{Where is the Vision space of the person who led a drug trafficking investigation?\\
      A. in the vicinity of the second man \quad B. in front of the woman\\
      C. in the left of the woman \quad D. in the background} \\

    \bottomrule
    \end{tabularx}
    \caption{Examples of single-modality and cross-modality coreference tasks. \textbf{In all cases, the model receives an audio-visual video and a textual biography as input}.}
  \label{tab:tasks_cases}
  \vspace{-0.6em}
\end{table*}

\begin{itemize}[itemsep=0.2em, topsep=0.2em, parsep=0pt, leftmargin=1em,
                label={\small$\blacktriangleright$}]
    \item \textbf{Audio:} Speech content and audio characteristics.
    \item \textbf{Vision:} Visual understanding, including actions, attributes, spatial reasoning, and counting.
    \item \textbf{Text:} Factual information from text, such as birthplace, career, and related attributes.
\end{itemize}

\subsection{Data Construction}
Our data construction pipeline consists of four stages: pre-processing, data annotation, modality alignment, and question–answer pair generation, as summarized in Figure~\ref{fig:pipeline}. We describe it below.

\subsubsection{Pre-Processing}
TVQA provides rich video and subtitle annotations, but their coverage and quality are uneven. Since our goal is cross-modal alignment that links each utterance to a speaker identity and synchronizes it with the corresponding visual evidence and text information, reliable human-labeled subtitles with accurate timestamps are essential. Therefore, we restrict our data to clips with manually annotated transcripts and discard segments that lack human subtitles or contain temporally discontinuous timestamps. Finally, we split each video into subtitle-aligned clips, which serve as the basic unit for the subsequent modality-specific annotation and question-answer pair construction.

\subsubsection{Data Annotation}
We annotate each video with visual, audio, and textual information. For the visual modality, we use Qwen2.5-VL-7B~\cite{bai2025qwen25-vl} to caption representative frames and then use Qwen3-8B~\cite{yang2025qwen3} to summarize them into clip-level descriptions, as well as to produce person-centric descriptions with external knowledge from subtitles (e.g., role cues explicitly mentioned in the subtitle). For audio, we treat the human-provided subtitles with speaker identities and timestamps as the ground truth for speech content. We additionally label basic voice characteristics (e.g., quality, pitch, emotion) using Qwen3-Omni-30B-A3B~\cite{Qwen3-Omni} through structured binary questions. Finally, we use Qwen3-8B~\cite{yang2025qwen3} to generate text biographies for people in the videos, including key attributes such as place of origin, important events, and production details. All descriptions are verified by cross-comparing annotations across different samples to ensure consistency and correctness. Prompts are provided in Table~\ref{app-tab:modality_annotation} and Table~\ref{app-tab:modality_annotation_db} in Appendix~\ref{sec:app-annotation-prompt}.

\subsubsection{Modality Alignment}
To align information across different modalities, we use visual features as the anchor. In particular, we treat each person as a bridge that links visual, audio, and textual information, thereby enabling tri-modal alignment. For visual–audio alignment, we use speaker names as anchors. The same name appearing in person-centric visual descriptions and human-labeled transcripts is used to associate the corresponding visual and audio segments. For visual–textual alignment, we integrate person-centric universal descriptions into textual biographies so that the biographies can be grounded in visual cues.

\begin{table}[t]
\centering
\scriptsize
\renewcommand{\baselinestretch}{1.1}
\begin{adjustbox}{max width=\textwidth, center}
\begin{tabular}{l l l}
\toprule
\multicolumn{1}{l}{\textbf{Model}} &
\multicolumn{1}{l}{\textbf{Base model}} &
\multicolumn{1}{l}{\textbf{Procedure}} \\
\midrule
Baichuan-Omni~\citeyearpar{li2024baichuan}   & Qwen2.5-7B & P+SFT \\
HumanomniV2~\citeyearpar{yang2025humanomniv2}     & Qwen2.5-Omni-7B & SFT+RL \\
M2-Omni~\citeyearpar{guo2025m2}        &  Llama3.1-8B & P+SFT  \\
Ming-Omni~\citeyearpar{ai2025ming}       & Ling-moe & P+SFT  \\
Minicpm-o~\citeyearpar{yao2024minicpm}         & Qwen2.5-7B & P+SFT+RL \\
Ola~\citeyearpar{liu2025ola}             & Qwen2.5-7B & P+SFT \\
Omni-r1~\citeyearpar{zhong2025omni}         & Qwen2.5-Omni-7B& SFT+RL \\
OmniVinci~\citeyearpar{ye2025omnivinci}       & Qwen2.5-7B & SFT \\
Qwen2.5-Omni~\citeyearpar{Qwen2.5-Omni}  & Qwen2.5-3B/7B & P+SFT \\
Qwen3-Omni-Cap~\citeyearpar{Qwen3-Omni}  & Qwen3-30B-A3B & P+SFT+RL \\
Qwen3-Omni-Ins~\citeyearpar{Qwen3-Omni}   & Qwen3-30B-A3B & P+SFT+RL \\
Qwen3-Omni-Thk~\citeyearpar{Qwen3-Omni}   & Qwen3-30B-A3B & P+SFT+RL \\
\bottomrule
\end{tabular}
\end{adjustbox}
\caption{Model statistics of different omni models. The P, SFT, and RL denote the pre-training, supervised fine-tuning, and reinforcement learning stages, respectively.}
\label{tab:model_stats}
\vspace{-1.5em}
\end{table}

\begin{table*}[t]
\centering
\small
\newlength{\metricw}
\settowidth{\metricw}{\textbf{Au$\rightarrow$Te}}

\newcolumntype{M}{>{\centering\arraybackslash}p{\metricw}} 

\renewcommand{\arraystretch}{1.05}
\setlength{\tabcolsep}{4.2pt}
\begin{adjustbox}{max width=0.96\textwidth, center}
\begin{tabular}{l *{9}{M} c}
\toprule
\multirow{2}{*}{\textbf{Model}} &
\multicolumn{3}{c}{\textbf{Audio-centric}} &
\multicolumn{3}{c}{\textbf{Text-centric}} &
\multicolumn{3}{c}{\textbf{Visual-centric}} &
\multirow{2}{*}{\textbf{Overall}} \\
\cmidrule(lr){2-4} \cmidrule(lr){5-7} \cmidrule(lr){8-10}
& \textbf{Au} & \textbf{Au$\rightarrow$Te} & \textbf{Au$\rightarrow$Vi}
& \textbf{Te} & \textbf{Te$\rightarrow$Au} & \textbf{Te$\rightarrow$Vi}
& \textbf{Vi} & \textbf{Vi$\rightarrow$Au} & \textbf{Vi$\rightarrow$Te} & \\
\midrule

Baichuan-Omni   & \underline{34.19} & 29.31 & 21.09 & \textbf{51.63} & 20.94 & 24.24 & 30.13 & 15.65 & 22.78 & 27.39 \\
HumanomniV2     & 28.75 & 39.08 & \underline{45.37} & 44.65 & 24.06 & 38.53 & \textbf{46.44} & 29.13 & 37.84 & 36.71 \\
M2-Omni         & 45.69 & 35.92 & 29.39 & \textbf{75.35} & 26.56 & 45.89 & \underline{56.49} & 33.04 & 51.35 & 42.83 \\
Ming-Omni       & 45.05 & 45.98 & 55.91 & \textbf{73.49} & 41.88 & 56.71 & \underline{64.44} & 37.39 & 49.81 & 51.38 \\
Minicpm         & 36.74 & 38.51 & 29.07 & \textbf{73.02} & 23.12 & 36.80 & 45.19 & 16.52 & \underline{47.49} & 37.48 \\
Ola             & 46.33 & 46.26 & 42.49 & \underline{49.30} & 39.38 & 46.75 & \textbf{54.81} & 29.57 & 37.07 & 43.52 \\
Omni-r1         & 49.20 & 46.84 & 45.69 & \textbf{83.26} & 38.44 & 54.11 & \underline{66.11} & 33.91 & 61.00 & 51.90 \\
OmniVinci       & 51.76 & 55.17 & \underline{56.23} & 55.35 & 39.69 & 47.62 & \textbf{59.83} & 32.17 & 44.79 & 49.39 \\
Qwen2.5-Omni-3B & 45.05 & 44.54 & 39.62 & \underline{57.21} & 39.06 & 50.65 & \textbf{63.18} & 33.04 & 44.02 & 45.62 \\
Qwen2.5-Omni-7B & 48.88 & 45.11 & 40.26 & \textbf{83.26} & 37.19 & 52.81 & \underline{64.44} & 33.91 & 59.85 & 50.36 \\
Qwen3-Omni-Cap  & \textbf{55.14} & 50.75 & 36.93 & 52.09 & 40.00 & 49.44 & \underline{53.47} & 38.26 & 50.12 & 47.19 \\
Qwen3-Omni-Ins  & \underline{60.58} & 53.05 & 44.92 & \textbf{64.65} & 44.69 & 55.06 & 58.91 & 43.48 & 56.29 & 53.10 \\
Qwen3-Omni-Thk  & 62.17 & 67.13 & 56.10 & \textbf{73.49} & 52.19 & 63.29 & 62.26 & 47.39 & \underline{69.42} & 61.29 \\
\midrule
\textbf{Average} &
46.89 & 45.97 & 41.77 & \textbf{64.36} & 35.94 & 47.84 & \underline{55.82} & 32.58 & 48.60 & 46.01 \\
\bottomrule
\end{tabular}
\end{adjustbox}
\caption{Performance of different models across modalities. Au, Te, and Vi denote Audio, Text, and Vision, respectively; X$\rightarrow$Y denotes source to target modality (e.g., Au$\rightarrow$Te: Audio to Text). In each row, the best is in \textbf{bold} and the second-best is \underline{underlined}.}
\label{tab:main_results}
\vspace{-1.0em}
\end{table*}

\subsubsection{Question--Answer Pair Construction}
We construct QA pairs in two steps: generation and verification. We use Qwen3-8B~\cite{yang2025qwen3} to generate questions based on the visual descriptions, subtitles, audio characteristics, and biographies, following the unified source-to-target format. For each question type, we enforce a fixed coreference transfer structure: locating the referent in a source modality and then selecting the corresponding information in a target modality. We then verify each candidate instance by checking (i) source-side uniqueness of the referent and (ii) target-side answerability from the annotations. Prompts are provided in Table~\ref{app-tab:qa_generation} in Appendix~\ref{sec:app-annotation-prompt}.

\subsection{Chain-of-Thought Rationales for Training}
For the training split, we provide structured rationales to support both analysis and training. Each rationale follows a four-step template aligned with our task format: (1) identify the source and target modalities from the question; (2) locate the referent in the source modality; (3) find the aligned referent in the target modality; and (4) extract the answer from the target modality. We generate these rationales with Qwen3-8B~\cite{yang2025qwen3} conditioned on the annotations and ground-truth answers, and verify them for consistency with both the evidence and the intended reasoning path. Prompts are provided in Table~\ref{app-tab:cot_generation} in Appendix~\ref{sec:app-annotation-prompt}.

\subsection{Quality Control and Human Verification}
We perform human verification to assess annotation quality and QA validity. For audio characteristics, we randomly sample 1{,}000 clips and re-label quality, pitch, and emotion, comparing human labels with those from the automatic pipeline and obtaining 93.1\% agreement. For video descriptions, we inspect 1{,}000 descriptions and obtain a correctness rate of 96.8\%. For textual biographies, we sample 1{,}000 cases and verify that there is no inappropriate or illegal content. For questions, we sample 1{,}000 instances and evaluate whether each question is answerable given the provided information, obtaining an answerability rate of 98.3\%. These results suggest that \textsc{CrossOmni} meets the quality requirements for reliable training and evaluation. More details are provided in Appendix~\ref{app:human_check}.

\section{Evaluation Experiment}
To examine the performance of coreference alignment across modalities, we evaluate Omni-LLMs on \textsc{CrossOmni}, which consists of single-modality questions and cross-modality counterparts under a coreference alignment task format.

\subsection{Experimental Details}
To comprehensively examine performance on cross-modal coreference alignment, we evaluate 13 mainstream Omni-LLMs with varying sizes, architectures, and training strategies. The model details are summarized in Table~\ref{tab:model_stats}. For all experiments, we use the prompt ``Answer this question based on the video and text information in A/B/C/D.'' and report accuracy as the evaluation metric.

\subsection{Main Evaluation Results}
The main results across different tasks are shown in Table~\ref{tab:main_results}. On average, cross-modality coreference tasks are about 21\% worse than single-modality coreference tasks. For almost all models, the top two results are obtained on single-modality coreference tasks, indicating that current Omni-LLMs have strong unimodal perception but do not consistently carry it over to cross-modal settings. Therefore, improving omni-modal reasoning calls for directly strengthening cross-modal coreference alignment, which appears to be a key missing component in current Omni-LLMs.

\begin{figure}[t]
    \centering

    \begin{subfigure}{0.95\linewidth}
        \centering
        \includegraphics[width=\linewidth]{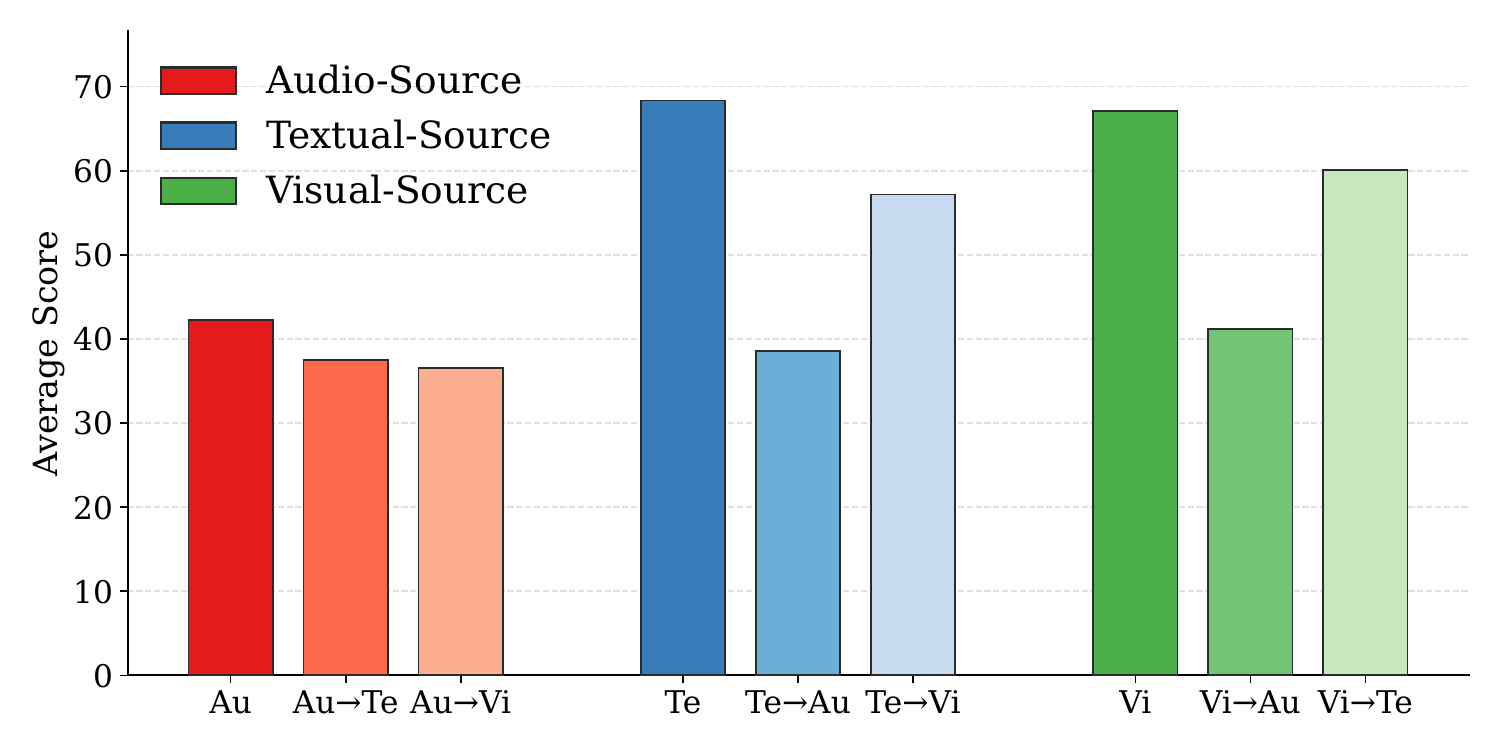}
        \caption{Average scores on fixed-source questions.}
        \label{fig:re_write}
        \vspace{-0.2em}
    \end{subfigure}

    \vspace{0.28em}

    \begin{subfigure}{0.95\linewidth}
        \centering
        \includegraphics[width=\linewidth]{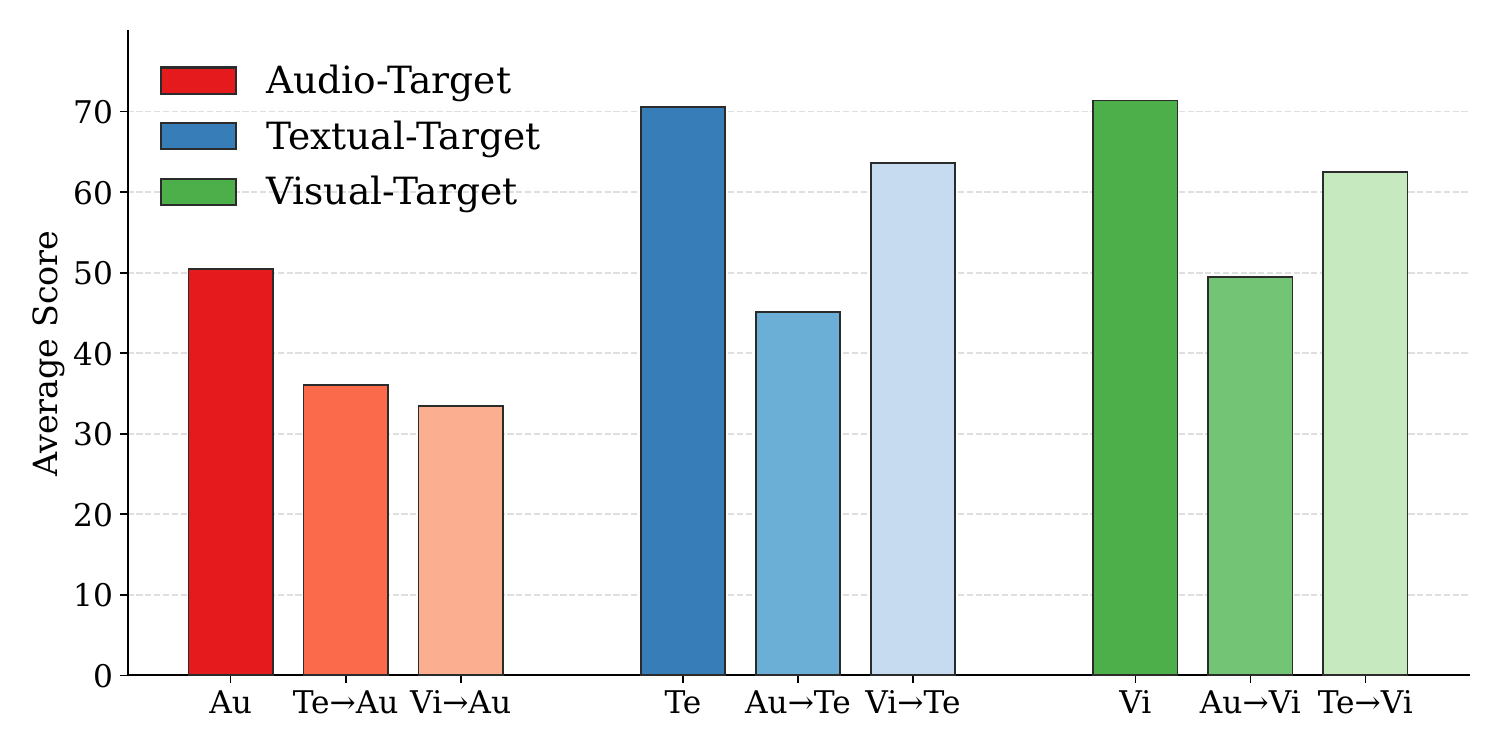}
        \caption{Average scores on fixed-target questions.}
        \label{fig:re_source}
        \vspace{-0.2em}
    \end{subfigure}

    \caption{Average performance on fixed-source and fixed-target cross-modal questions.}

    \label{fig:re_fixed}
    \vspace{-1.5em}
\end{figure}

\subsection{Controlled Cross-modal Experiments}
To disentangle the effect of information content from that of cross-modal coreference alignment, we construct two controlled settings. For fixed-source questions, we select 150 questions from each single-modality coreference task and convert them into cross-modality coreference questions by changing only the target modality while keeping the source information fixed. For fixed-target questions, we follow a similar process but keep the target modality fixed and vary the source modality. As shown in Figure~\ref{fig:re_source} and Figure~\ref{fig:re_write}, performance consistently drops once the source modality differs from the target modality, even though the information in the fixed modality remains unchanged. This indicates that current models struggle to identify the correct referent in the source modality and transfer it reliably to the target modality, further demonstrating the necessity of improving cross-modal coreference alignment.

\section{Analysis and Improvements}
To better understand and improve cross-modal coreference alignment in Omni-LLMs, we conduct a systematic analysis of model failures and develop targeted remedies. We first diagnose why models perform well when evidence is contained within a single modality but fail when the answer requires transferring a referent across modalities.  Our analyses indicate that a central bottleneck is the lack of explicit coreference-aware thinking patterns for cross-modality reasoning.  We then present two complementary approaches to induce such procedures: a training-free In-Context Learning method and training-based framework via SFT and GRPO. Experimental results show that enhancing it effectively improves overall reasoning performance.

\subsection{Diagnosing Failures Cause}
\label{sec:further_diss}

\begin{figure*}[t]
    \centering
    \includegraphics[width=0.92\linewidth]{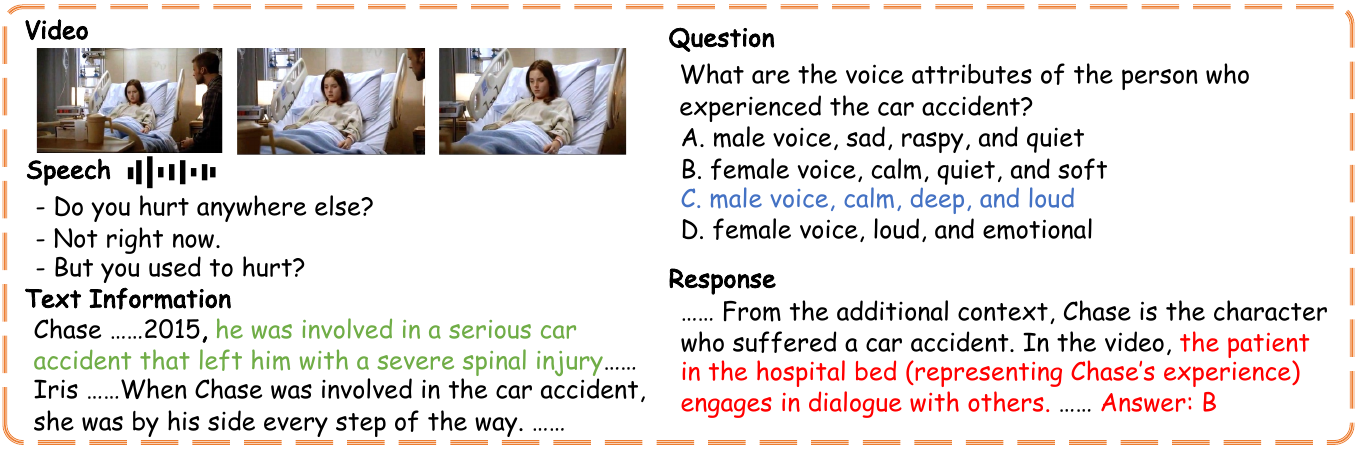}
    \caption{CoT case study on text$\rightarrow$audio task. The \textcolor{cot-green}{green} denotes important information, the \textcolor{cot-blue}{blue} denotes the right choice, and the \textcolor{cot-red}{red} denotes the error rationales. The model fails to establish the correct bridge between the textual information (the person who experienced the car accident is a man) and the visual information (the man is sitting on the chair), and therefore cannot answer the question correctly. The full case is provide in Figure~\ref{fig:cot_case} in Appendix~\ref{app:case_study}.}
    \label{fig:cot_case_short}
    \vspace{-1em}
\end{figure*}

\begin{figure}[t]
    \centering
    \includegraphics[width=0.6\linewidth]{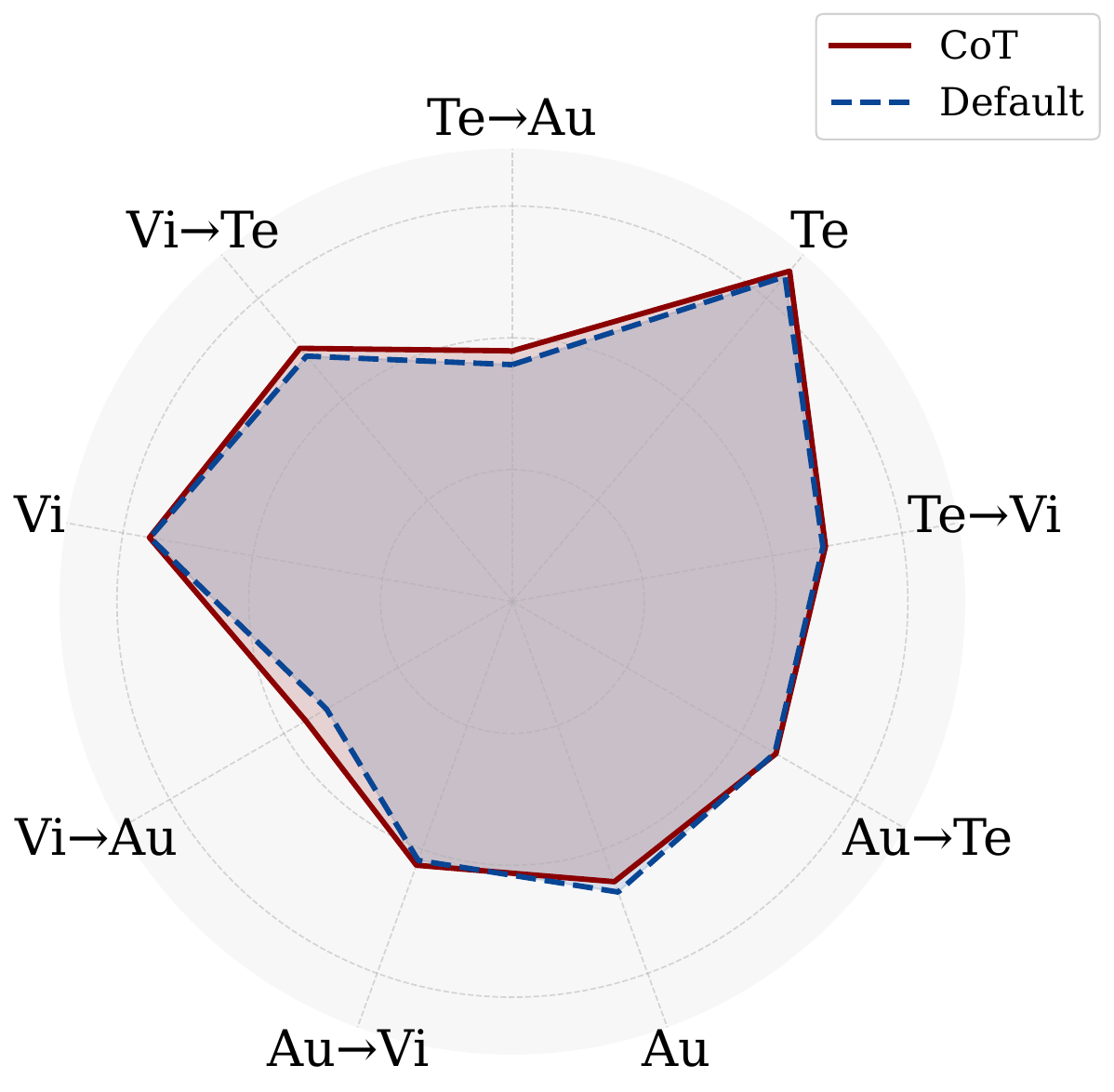}
    \caption{Accuracy comparison of 13 models under default and Chain-of-Thought prompting.}
    \label{fig:cot}
    \vspace{-1 em}
\end{figure}
To pinpoint the sources of cross-modal coreference failures, we adopt Chain-of-Thought (CoT) prompting to make the model explicitly reveal its coreference alignment process by prepending ``Let’s think step by step.'' to the default prompt. This setting allows us to inspect the resulting reasoning rationales of the whole process for systematic errors.

Importantly, we find that models rarely establish explicit and correct links between entities across modalities, as shown in Figure~\ref{fig:cot_case_short}, indicating that current models lack coreference-aware thinking patterns. Furthermore, as shown in Figure~\ref{fig:cot} and Table~\ref{tab:method_comparison}, Chain-of-Thought prompting brings only modest gains, with a relative improvement of 1.7\% overall. 
These limited improvements suggest that current models lack sufficiently strong cross-modal coreference thinking capabilities on their own, and that merely eliciting more detailed thinking is far from sufficient to close the gap. Consequently, improving cross-modal coreference alignment requires explicitly shaping coreference-aware thinking patterns for reference alignment.

\subsection{Improving Cross-Modal Coreference}
\label{sec:methods}
Building on the above experiments, which show that cross-modality coreference performance lags far behind single-modality coreference and model-internal thinking patterns yield only marginal gains, we regard the lack of coreference-aware thinking patterns as the main bottleneck. Therefore, we explore two approaches to enhance the coreference-aware thinking patterns for modality coreference alignment: a training-free approach based on In-Context Learning (ICL), and training-based approaches using Supervised Fine-Tuning (SFT) and Group Relative Policy Optimization (GRPO).

\subsubsection{Training-free: In-Context Learning}
Given that existing models can correctly extract information within a single modality but struggle to align it across modalities, we design a dedicated prompt to guide step-by-step cross-modal coreference alignment via In-Context Learning (ICL), as shown in Table~\ref{app-tab:icl_prompt} in Appendix~\ref{sec:app-icl-prompt}. As reported in Table~\ref{tab:method_comparison}, applying ICL yields substantial performance gains, particularly on cross-modal coreference tasks, with a 21\% relative improvement. The performance gap between single-modality and cross-modality coreference shrinks from 24.37\% to 11.01\% under ICL. 
These results are consistent with our diagnosis that the primary limitation in cross-modal alignment lies in coreference-aware thinking patterns, indicating that future work should focus on explicitly learning and refining these thinking patterns.

\subsubsection{Training-based: SFT and GRPO}
\paragraph{Training Details.}
We use both SFT-based and GRPO-based training methods to enhance cross-modal coreference capability. For SFT, we perform LoRA fine-tuning and then apply GRPO on top of the SFT checkpoint using 982 training instances balanced across modalities, with Qwen3-Next-80B-A3B-Instruct~\cite{yang2025qwen3} as the reward model. More details are provided in Appendix~\ref{app:training_detail}.

\paragraph{Experimental Results.}

\begin{table}[t]
    \centering
    \small
    \setlength{\tabcolsep}{6pt}
    \renewcommand{\arraystretch}{1.12}

    \begin{adjustbox}{max width=0.49\textwidth, center}
    \begin{tabular}{@{}lccc@{}}
        \toprule
        \textbf{Method} & \textbf{Single-modality} & \textbf{Cross-modality} & \textbf{Overall} \\
        \midrule
        Default
        & 55.69
        & 42.12
        & 46.01 \\
        CoT
        & 55.52\,{\textcolor{myblue}{\scriptsize($-0.30\%$)}}
        & 43.56\,{\textcolor{myred}{\scriptsize($+3.42\%$)}}
        & 46.83\,{\textcolor{myred}{\scriptsize($+1.78\%$)}} \\
        \addlinespace[1pt]
        \textbf{ICL}
        & 57.29\,{\textcolor{myred}{\scriptsize($+2.87\%$)}}
        & \phantom{0}50.98\,{\textcolor{myred}{\scriptsize($+21.04\%$)}}
        & \phantom{0}52.63\,{\textcolor{myred}{\scriptsize($+14.39\%$)}} \\
        \bottomrule
    \end{tabular}
    \end{adjustbox}
    \caption{Average performance of 13 models with CoT and ICL on \textsc{CrossOmni}.}
    \label{tab:method_comparison}
    \vspace{-1.5em}
\end{table}

\begin{table*}[htbp]
\centering
\renewcommand{\baselinestretch}{1.11}

\settowidth{\metricw}{\textbf{Au$\rightarrow$Te}}

\sisetup{
  table-number-alignment = center,
  table-column-width = \metricw
}

\begin{adjustbox}{max width=0.98\textwidth, center}
\begin{tabular}{ll *{9}{S[table-format=2.2]} c}
\toprule
\multicolumn{1}{c}{\multirow{2}{*}{\textbf{Model}}} &
\multicolumn{1}{c}{\multirow{2}{*}{\textbf{Method}}} &
\multicolumn{3}{c}{\textbf{Audio-centric}} &
\multicolumn{3}{c}{\textbf{Text-centric}} &
\multicolumn{3}{c}{\textbf{Visual-centric}} &
\multicolumn{1}{c}{\multirow{2}{*}{\textbf{Overall}}}\\
\cmidrule(lr){3-5} \cmidrule(lr){6-8} \cmidrule(lr){9-11}
& & {\textbf{Au}} & {\textbf{Au$\rightarrow$Te}} & {\textbf{Au$\rightarrow$Vi}}
& {\textbf{Te}} & {\textbf{Te$\rightarrow$Au}} & {\textbf{Te$\rightarrow$Vi}}
& {\textbf{Vi}} & {\textbf{Vi$\rightarrow$Au}} & {\textbf{Vi$\rightarrow$Te}}
& \\
\midrule

\multirow{4}{*}{Qwen2.5-Omni-3B}
& Default       & 45.05 & 44.54 & 39.62 & 57.21 & 39.06 & 50.65 & 63.18 & 33.04 & 44.02 & 45.62\\
& CoT           & 44.09 & 42.53 & 42.81 & 50.23 & 37.81 & 48.92 & 66.11 & 33.91 & 39.38 & 44.57\,{\textcolor{myblue}{\scriptsize (-2.30\%)}} \\
& \textbf{SFT}      & 39.62 & 50.28 & 52.72 & 76.74 & 38.75 & 54.54 & 58.99 & 40.00 & 51.35 & \phantom{0}50.45\,{\textcolor{myred}{\scriptsize (+10.59\%)}} \\
& \textbf{GRPO}     & 46.33 & 60.34 & 56.55 & 87.44 & 46.25 & 62.34 & 69.04 & 43.48 & 65.25 & \phantom{0}58.59\,{\textcolor{myred}{\scriptsize (+28.43\%)}} \\
\midrule

\multirow{4}{*}{Qwen2.5-Omni-7B}
& Default       & 48.88 & 45.11 & 40.26 & 83.26 & 37.19 & 52.81 & 64.44 & 33.91 & 59.85 & 50.36\\
& CoT           & 43.77 & 46.26 & 37.70 & 83.26 & 36.56 & 55.84 & 63.18 & 37.39 & 58.69 & 49.84\,{\textcolor{myblue}{\scriptsize (-1.03\%)}} \\
& \textbf{SFT}     & 50.48 & 56.90 & 54.63 & 80.46 & 47.19 & 60.61 & 59.83 & 45.65 & 59.83 & \phantom{0}56.65\,{\textcolor{myred}{\scriptsize (+12.49\%)}} \\
& \textbf{GRPO}     & 57.19 & 62.93 & 61.02 & 88.84 & 51.56 & 63.64 & 67.36 & 56.48 & 64.86 & \phantom{0}62.56\,{\textcolor{myred}{\scriptsize (+24.23\%)}} \\
\bottomrule
\end{tabular}
\end{adjustbox}

\caption{Performance of Qwen2.5-Omni models across modalities and training methods.
Au, Te, and Vi denote Audio, Text, and Visual, respectively. X$\rightarrow$Y denotes source-to-target modality (e.g., Au$\rightarrow$Te: Audio-to-Text).}
\label{tab:training}
\vspace{-0.8em}
\end{table*}

To further enhance coreference-aware thinking patterns for cross-modal alignment, we conduct experiments on Qwen2.5-Omni-3B and Qwen2.5-Omni-7B using SFT and GRPO training methods. To establish the coreference-aware thinking patterns, we apply CoT supervision during SFT and introduce a thinking-pattern reward during GRPO. Specifically, the GRPO objective combines rewards for output format, reasoning accuracy, contextual consistency, and logical coherence. The details are provided in the Table~\ref{app-tab:grpo_prompt} in Appendix~\ref{sec:app-GRPO-prompt}. As shown in Table~\ref{tab:training}, both SFT and GRPO yield substantial performance gains, with GRPO achieving the largest improvements, particularly on cross-modal coreference tasks. We further present two representative cases in Figure~\ref{fig:case1} and Figure~\ref{fig:case2} in Appendix~\ref{app:case_study}, which illustrate the effectiveness of these patterns. Overall, these results demonstrate that explicitly training coreference-aware thinking patterns is crucial for improving cross-modal performance.

\paragraph{Necessity of Explicit Coreference-aware Thinking Patterns.}

\begin{table}[t]
    \centering
    \small
    \setlength{\tabcolsep}{4.5pt}
    \renewcommand{\arraystretch}{1.12}

    \begin{adjustbox}{max width=0.49\textwidth, center}
    \begin{tabular}{@{}lcccccc@{}}
        \toprule
        \multirow{1}{*}{\textbf{Method}} 
        & \multicolumn{1}{c}{\textbf{Single-modality}} 
        & \multicolumn{1}{c}{\textbf{Cross-modality}} 
        & \multicolumn{1}{c}{\textbf{Overall}} \\
        \midrule

        \multicolumn{7}{c}{\textbf{Qwen2.5-Omni-3B}} \\
        \midrule
        Default       
        & 55.15 
        & 41.82 
        & 45.62 \\
        \hdashline

        \textbf{SFT}
        & 58.45\,{\textcolor{myred}{\scriptsize($+5.98\%$)}}
        & \phantom{0}47.94\,{\textcolor{myred}{\scriptsize($+14.63\%$)}}
        & \phantom{0}50.45\,{\textcolor{myred}{\scriptsize($+10.59\%$)}} \\

        SFT w/o CoT
        & 50.15\,{\textcolor{myblue}{\scriptsize($-9.06\%$)}}
        & 42.55\,{\textcolor{myred}{\scriptsize($+1.75\%$)}}
        & 45.22\,{\textcolor{myblue}{\scriptsize($-0.88\%$)}} \\

        \addlinespace[2pt]
        \midrule
        \multicolumn{7}{c}{\textbf{Qwen2.5-Omni-7B}} \\
        \midrule
        Default       
        & 65.52
        & 44.86
        & 50.36\\
        \hdashline

        \textbf{SFT}
        & 63.59\,{\textcolor{myblue}{\scriptsize($-2.95\%$)}}
        & \phantom{0}54.39\,{\textcolor{myred}{\scriptsize($+21.24\%$)}}
        & \phantom{0}56.65\,{\textcolor{myred}{\scriptsize($+12.49\%$)}} \\

        SFT w/o CoT
        & \phantom{0}52.42\,{\textcolor{myblue}{\scriptsize($-19.99\%$)}}
        & 46.30\,{\textcolor{myred}{\scriptsize($+3.21\%$)}}
        & 48.70\,{\textcolor{myblue}{\scriptsize($-3.30\%$)}} \\
        \bottomrule
    \end{tabular}
    \end{adjustbox}
    \caption{Ablation study of SFT on Qwen2.5-Omni.}
    \label{tab:sft-abl}
\end{table}

\begin{table}[t]
    \centering
    \small
    \setlength{\tabcolsep}{6pt}
    \renewcommand{\arraystretch}{1.12}

    \begin{adjustbox}{max width=0.49\textwidth, center}
    \begin{tabular}{@{}lccc@{}}
        \toprule
        \textbf{Method} & \textbf{Single-modality} & \textbf{Cross-modality} & \textbf{Overall} \\
        \midrule

        \multicolumn{4}{c}{\textbf{Qwen2.5-Omni-3B}} \\
        \midrule
        SFT      
        & 58.45
        & 47.94
        & 50.45 \\
        \hdashline

        \textbf{GRPO}
        & \phantom{0}67.60\,{\textcolor{myred}{\scriptsize($+15.65\%$)}}
        & \phantom{0}55.70\,{\textcolor{myred}{\scriptsize($+16.19\%$)}}
        & \phantom{0}58.59\,{\textcolor{myred}{\scriptsize($+16.13\%$)}} \\

        GRPO w/o T-R
        & 58.98\,{\textcolor{myred}{\scriptsize($+0.91\%$)}}
        & 50.03\,{\textcolor{myred}{\scriptsize($+4.36\%$)}}
        & 52.67\,{\textcolor{myred}{\scriptsize($+4.40\%$)}} \\

        \addlinespace[2pt]
        \midrule
        \multicolumn{4}{c}{\textbf{Qwen2.5-Omni-7B}} \\
        \midrule
        SFT       
        & 63.59
        & 54.39
        & 56.65 \\
        \hdashline

        \textbf{GRPO}
        & \phantom{0}71.13\,{\textcolor{myred}{\scriptsize($+11.86\%$)}}
        & 59.58\,{\textcolor{myred}{\scriptsize($+9.54\%$)}}
        & \phantom{0}62.56\,{\textcolor{myred}{\scriptsize($+10.43\%$)}} \\

        GRPO w/o T-R
        & 67.60\,{\textcolor{myred}{\scriptsize($+6.31\%$)}}
        & 57.55\,{\textcolor{myred}{\scriptsize($+5.81\%$)}}
        & 59.72\,{\textcolor{myred}{\scriptsize($+5.42\%$)}} \\
        \bottomrule
    \end{tabular}
    \end{adjustbox}
    \caption{Ablation study of GRPO on Qwen2.5-Omni. T-R denotes the thinking-pattern reward.}
    \label{tab:GRPO-abl}
    \vspace{-0.9em}
\end{table}

To further isolate the contribution of the coreference-aware thinking patterns in the SFT and GRPO stages, we conduct ablations that remove them from each training process. Specifically, we perform SFT without CoT supervision and GRPO without thinking patterns rewards. Tables~\ref{tab:sft-abl} and~\ref{tab:GRPO-abl} report consistent performance drops under both ablations, showing that CoT supervision and thinking-pattern rewards are effective for establishing coreference-aware reasoning. These results further indicate that explicitly modeling coreference-aware thinking patterns is crucial for effective cross-modal reasoning.

\paragraph{Generalization to OOD Benchmarks.}

\begin{table}[t]
    \centering
    \small
    \setlength{\tabcolsep}{4.2pt}
    \renewcommand{\arraystretch}{1.2}

    \begin{adjustbox}{max width=0.49\textwidth, center}
    \begin{tabular}{@{}lcccc@{}}
        \toprule
        \multirow{1}{*}{\textbf{Method}} 
        & \multicolumn{1}{c}{\textbf{Daily-Omni}} 
        & \multicolumn{1}{c}{\textbf{WorldSense}} 
        & \multicolumn{1}{c}{\textbf{AVUT}}
        & \multicolumn{1}{c}{\textbf{Average}} \\
        \midrule

        \multicolumn{5}{c}{\textbf{Qwen2.5-Omni-3B}} \\
        \midrule
        Default       
        & 40.52
        & 43.50
        & 48.10
        & 44.04\\
        \textbf{SFT}       
        & 47.45\,{\textcolor{myred}{\scriptsize($+17.10\%$)}}
        & 39.24\,{\textcolor{myblue}{\scriptsize($-9.79\%$)}}
        & 48.79\,{\textcolor{myred}{\scriptsize($+1.43\%$)}}
        & 45.16\,{\textcolor{myred}{\scriptsize($+2.54\%$)}} \\
        \textbf{GRPO}    
        & 54.55\,{\textcolor{myred}{\scriptsize($+34.62\%$)}}
        & 44.45\,{\textcolor{myred}{\scriptsize($+2.18\%$)}}
        & \phantom{0}57.79\,{\textcolor{myred}{\scriptsize($+20.14\%$)}}
        & \phantom{0}52.26\,{\textcolor{myred}{\scriptsize($+18.67\%$)}} \\

        \addlinespace[2pt]
        \midrule
        \multicolumn{5}{c}{\textbf{Qwen2.5-Omni-7B}} \\
        \midrule
        Default       
        & 47.45
        & 45.40
        & 48.90
        & 47.25 \\ 
        \textbf{SFT}       
        & 55.97\,{\textcolor{myred}{\scriptsize($+17.96\%$)}}
        & 41.36\,{\textcolor{myblue}{\scriptsize($-8.90\%$)}}
        & 49.77\,{\textcolor{myred}{\scriptsize($+1.78\%$)}}
        & 49.03\,{\textcolor{myred}{\scriptsize($+3.77\%$)}} \\
        \textbf{GRPO}       
        & 57.89\,{\textcolor{myred}{\scriptsize($+22.00\%$)}}
        & 49.08\,{\textcolor{myred}{\scriptsize($+8.11\%$)}}
        & \phantom{0}57.96\,{\textcolor{myred}{\scriptsize($+18.52\%$)}}
        & \phantom{0}54.98\,{\textcolor{myred}{\scriptsize($+16.35\%$)}} \\
        \bottomrule
    \end{tabular}
    \end{adjustbox}
    \caption{Performance on Daily-Omni, WorldSense, and AVUT dataset. The details about these datasets are provided in Appendix~\ref{app:dataset}.}
    \label{tab:ood}
    \vspace{-0.6em}
\end{table}
 
To examine whether the cross-modal coreference alignment ability learned on \textsc{CrossOmni} generalizes, we evaluate the trained models on three out-of-distribution (OOD) benchmarks for collaborative reasoning and cross-modal alignment.
As reported in Table~\ref{tab:ood}, both training strategies achieve consistent gains on almost all benchmarks, with especially great improvements on collaborative reasoning tasks in the Daily-Omni and WorldSense benchmarks, highlighting the importance of coreference-aware reasoning for both cross-modal alignment and omni-modal collaborative reasoning.
On WorldSense, SFT alone slightly degrades performance because of a mismatch in video scenarios between WorldSense and \textsc{CrossOmni}, whereas subsequent GRPO not only recovers this drop but also yields clear gains, indicating that GRPO is essential for coreference-aware training, as it learns robust cross-modal coreference alignment instead of overfitting to the \textsc{CrossOmni} distribution.
Overall, these results demonstrate that cross-modal coreference alignment learned from our dataset generalizes to OOD settings and constitutes a critical capability for advancing Omni-LLMs.

\section{Conclusions}
To advance the reasoning capabilities of Omni-LLMs, we focus on cross-modal coreference alignment and introduce the \textsc{CrossOmni} dataset for evaluating and training this capability. Across 13 Omni-LLMs, we observe that unimodal competence does not directly translate into reliable cross-modal evidence transfer. Our analyses suggest that a key failure factor is the absence of coreference-aware thinking patterns for aligning referents across modalities. Motivated by this, we propose a training-free In-Context Learning method and a training-based SFT+GRPO framework to induce such patterns, substantially improving performance on \textsc{CrossOmni} and generalizing to broader collaborative reasoning tasks. These results highlight cross-modal coreference as a crucial missing piece for improving the reasoning capabilities of Omni-LLMs and provide a concrete pathway toward more reliable omni-modal understanding.

\clearpage
\section*{Limitations}
There are several limitations in this work. First, performance on audio-based tasks lags behind that on text-based and visual-based tasks. We attribute this degradation to weaker auditory perception and reasoning, especially in multi-speaker scenarios where the model must disentangle overlapping speech, identify speakers, and track their mentions over time. However, we have not yet conducted a thorough analysis of the underlying factors or effective remedies, and we plan to investigate this gap in our next work through more fine-grained error analysis and controlled ablations. Second, our annotation pipeline currently relies on human-provided subtitles, which limits end-to-end automation and scalability and makes it difficult to extend the benchmark to truly large-scale real-world corpora.

\section*{Ethical Considerations}
All videos in our dataset are sampled from TVQA, which is publicly available for academic use. We follow its license terms, use the data solely for research, and do not redistribute raw video content. All models in our pipeline and evaluation are open-source, and our processing does not introduce additional personally identifiable information beyond what is present in TVQA, we only release derived annotations. College students provided human ratings without collecting personal characteristics or identifiers. We also used large language models as writing assistants for editing and polishing the paper. Overall, we believe our work poses minimal additional ethical or privacy risks beyond those already present in the underlying TVQA dataset.

\newpage
\bibliography{custom}

\clearpage
\appendix
\section{Related Works}
\paragraph{Omni-LLMs}
Omni-LLMs are designed to jointly process textual information, auditory features, and visual content, and to generate accurate responses based on integrated omni-modal understanding and reasoning~\cite{liu-etal-2024-ce,ye2025omnivinci,team2025longcat,11471045}. However, compared with holistic multi-modal perception, effectively leveraging omni-modal information in complex scenarios remains a central challenge~\cite{liu2024m2k,ai2025mingflash,10447469}. For example, Baichuan-Omni~\cite{li2024baichuan} adopts a specialized audio–visual alignment procedure via SFT, and Qwen3-Omni~\cite{Qwen3-Omni} applies GRPO to enhance collaboration among the three modalities. These models, however, mainly focus on aggregating multiple modalities and largely ignore fine-grained cross-modal alignment, which severely limits their overall reasoning performance. To bridge this gap, we formulate cross-modal coreference alignment as a core problem, requiring models to establish correct correspondences for the same referent across modalities and offering a new perspective for improving the overall performance of Omni-LLMs.

\paragraph{Omni-Modal Dataset} 
The omni-modal dataset aims to enhance the capabilities of Omni-LLMs across diverse modalities. From a broader perspective, however, mainstream evaluation of Omni-LLMs still predominantly relies on single-modality benchmarks~\cite{rein2024gpqa,wang2025mmsu,liu2025vocalbench}, such as MMMU~\cite{yue2024mmmu} for image understanding and MMLU~\cite{wang2024mmlu} for text comprehension. In parallel, many works have been proposed to probe unimodal reasoning capabilities~\cite{yizhiomnibench,xing2025echoink}. For example, DailyOmni~\cite{zhou2025dailyomni} targets audio and visual reasoning in everyday scenarios, and WorldSense~\cite{hong2025worldsense} focuses on assessing collaborative understanding and reasoning over omni-modal inputs.
While these datasets are effective for measuring overall unimodal or holistic omni-modal performance, they typically do not isolate the key step of cross-modality coreference alignment, which limits their ability to diagnose why unimodal success does not translate into reliable omni-modal reasoning. To fill this gap, we introduce \textsc{CrossOmni}, a dataset that explicitly targets source-to-target coreference alignment, enabling systematic analysis of cross-modal coreference failures, as well as providing training data to improve cross-modal coreference alignment.

\section{Experimental Details}
\subsection{Task Classification Details}
We construct three single-modality coreference tasks and six cross-modality coreference tasks. For cross-modality coreference tasks that involve audio and text, we use the visual modality as an anchor to ground the referent and connect modalities through person-centric alignment. The details of each task are as follows:
\label{app:task_detail}
\paragraph{Single-Modality Coreference Tasks.}
\begin{itemize}[itemsep=0.2em, topsep=0.2em, parsep=0pt, leftmargin=1.5em,
                label={\small$\blacktriangleright$}]
    \item \textbf{Audio-only (Au).} Understanding speech content and audio characteristics.
    \item \textbf{Visual-only (Vi).} Visual understanding, including actions, attributes, spatial reasoning, and counting.
    \item \textbf{Text-only (Te).} Factual information from text, such as birthplace, career, education, and related attributes.
\end{itemize}

\paragraph{Cross-Modality Coreference Tasks.}
\begin{itemize}[itemsep=0.2em, topsep=0.2em, parsep=0pt, leftmargin=1.5em,
                label={\small$\blacktriangleright$}]
    \item \textbf{Audio$\rightarrow$Vision (Au$\rightarrow$Vi).} Identify the person from the audio and answer about their visual information.
    \item \textbf{Audio$\rightarrow$Text (Au$\rightarrow$Te).} Identify the person from audio, ground via a visual anchor, then answer textual facts.
    \item \textbf{Vision$\rightarrow$Audio (Vi$\rightarrow$Au).} Identify the person from visual cues and answer about speech transcription or audio characteristics.
    \item \textbf{Vision$\rightarrow$Text (Vi$\rightarrow$Te).} Identify the person from visual cues and answer textual facts.
    \item \textbf{Text$\rightarrow$Audio (Te$\rightarrow$Au).} Identify the person from the text, ground via a visual anchor, then answer speech transcription or audio characteristics.
    \item \textbf{Text$\rightarrow$Vision (Te$\rightarrow$Vi).} Identify the person from the text and answer about the visual information.
\end{itemize}
\subsection{Training Details}
\label{app:training_detail}
We use both SFT and GRPO training to enhance cross-modal coreference capability. For SFT, we perform LoRA fine-tuning with rank 16 and $\alpha=32$, using a learning rate of $1\times10^{-5}$, a batch size of 32, and 1 training epoch. For GRPO, we initialize the policy from the SFT checkpoint and sample 982 instances from the training set, ensuring that each modality has a similar number of examples. We use a learning rate of $5\times10^{-6}$, a batch size of 16, a rollout number of 4, and 1 GRPO epoch, with Qwen3-Next-80B-A3B-Instruct~\cite{yang2025qwen3} as the reward model. 
\subsection{Datasets on Ablation Study}
\label{app:dataset}
We conduct ablation studies on three omni datasets, and the detailed descriptions are as follows:
\paragraph{Daily-Omni:}
Daily-Omni~\cite{zhou2025dailyomni} is an audio-visual multiple-choice QA benchmark targeting temporally aligned cross-modal reasoning in daily-life scenarios. It contains 684 videos and 1,197 QA pairs, with clips segmented into 30\,s or 60\,s to probe temporal understanding under different context lengths. The questions are grouped into six task types, all designed to require integrating audio and visual evidence rather than relying on a single modality.

\paragraph{WorldSense:}
WorldSense~\cite{hong2025worldsense} is a real-world omni-modal video understanding benchmark that evaluates models with jointly provided video and audio, emphasizing strong audio-visual coupling such that removing either modality degrades answerability. It contains 1,662 audio-visual synchronized videos with 3,172 multiple-choice QA pairs spanning 26 distinct tasks.

\paragraph{AVUT:}
AVUT~\cite{yang2025avut} is an audio-centric video understanding benchmark designed to evaluate auditory understanding and audio-visual interactions, while containing sufficient audio-visual cross-modality tasks. It includes 2,662 YouTube videos covering 18 audio-centric domains and 11,609 QA pairs over 8 tasks, with an average duration of 67.8 seconds. We choose the AV-Human as the test set.

\subsection{Human Verification}
\label{app:human_check}
We provide details of the human verification procedure used in data construction. Five college students are recruited to verify the quality of the annotations. 
For audio characteristics, each clip is annotated with a set of labels (e.g., voice quality, pitch, emotion), and we treat human labels as the reference. 
We treat a label as an error if it is predicted by the auto-labeling system but absent from the human annotations, and we measure agreement as the proportion of automatically predicted labels that are also present in the human-labeled set.
For video descriptions, we check whether the generated descriptions are inconsistent with the video content using a binary label (True/False). 
For textual biographies, we verify whether there is any inappropriate or illegal content, again with a binary label. 
For questions, we verify each question against the answer that is supported by the evidence, also using a binary label.
\subsection{Further Discussion on Main Results}
We further analyze the results in Table~\ref{tab:main_results} and obtain several conclusions as follows:
\begin{enumerate}
    \item Model size helps but is not sufficient: We observe that a smaller model can outperform a larger one (e.g., Qwen2.5-Omni-3B vs. Ola-7B in our table), indicating that data/recipe matters as much as parameter count.
    \item Architecture, especially the audio encoder, is critical: models using stronger Whisper variants tend to be markedly better, consistent with MiniCPM-o building on Whisper-medium while Qwen2.5-Omni is designed with Whisper-large-v3.
    \item Training paradigm matters: ``thinking''-enhanced omni variants (e.g., Qwen3-Omni-Thinking) show stronger reasoning and can improve cross-modal alignment.
    \item Data scaling is essential: expanding alignment-focused data consistently improves both SFT and GRPO in our experiments, highlighting the importance of sufficient, well-curated training signals.
\end{enumerate}
Overall, for cross-modal alignment, strong results typically require adequate model capacity, a high-quality audio front-end, explicit reasoning training, and sufficient training

\subsection{Detailed Results on CoT and ICL}The detailed results of Chain-of-Thought prompting and In-Context Learning are reported in Table~\ref{tab:cot_results} and Table~\ref{tab:icl_results}. Most models benefit from In-Context Learning, with only two exceptions: Qwen2.5-Omni-3B shows degraded performance, potentially due to limited instruction-following ability, and Qwen3-Omni-Thinking also underperforms, likely because it relies more heavily on its internal thinking patterns rather than externally provided demonstrations. The consistent advantage of in-context exemplars over a model's native ``thinking'' behavior further suggests that the primary bottleneck of current omni-LLMs lies in the lack of effective coreference-aware thinking patterns for cross-modality coreference alignment.

\section{Case Study}
\label{app:case_study}
\begin{figure*}
    \centering
    \includegraphics[width=0.9\linewidth]{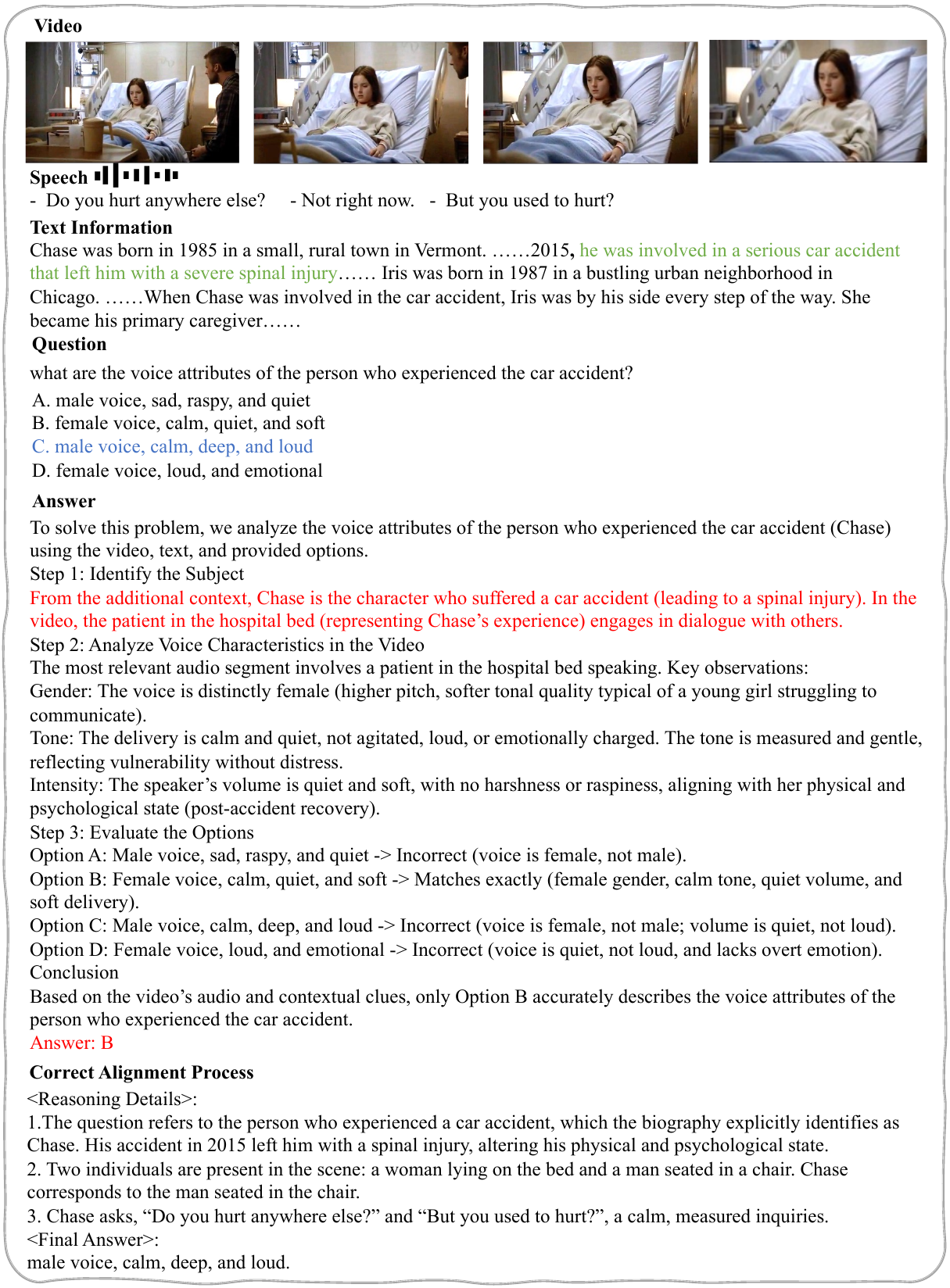}
    \caption{Case study on text$\rightarrow$audio task via Qwen3-Omni-30B-A3B-Thinking via CoT. The \textcolor{cot-green}{green} denotes important information, the \textcolor{cot-blue}{blue} denotes the right choice, and the \textcolor{cot-red}{red} denotes the error rationales. The model fails to establish the correct bridge between the textual information (the person who experienced the car accident is a man) and the visual information (the man is sitting on the chair), and therefore cannot answer the question correctly. The correct process should first identify Chase in the text and extract that this person is a man, then locate his position in the video, and finally link his speech in the audio.}
    \label{fig:cot_case}
\end{figure*}

\begin{figure*}
    \centering
    \includegraphics[width=0.9\linewidth]{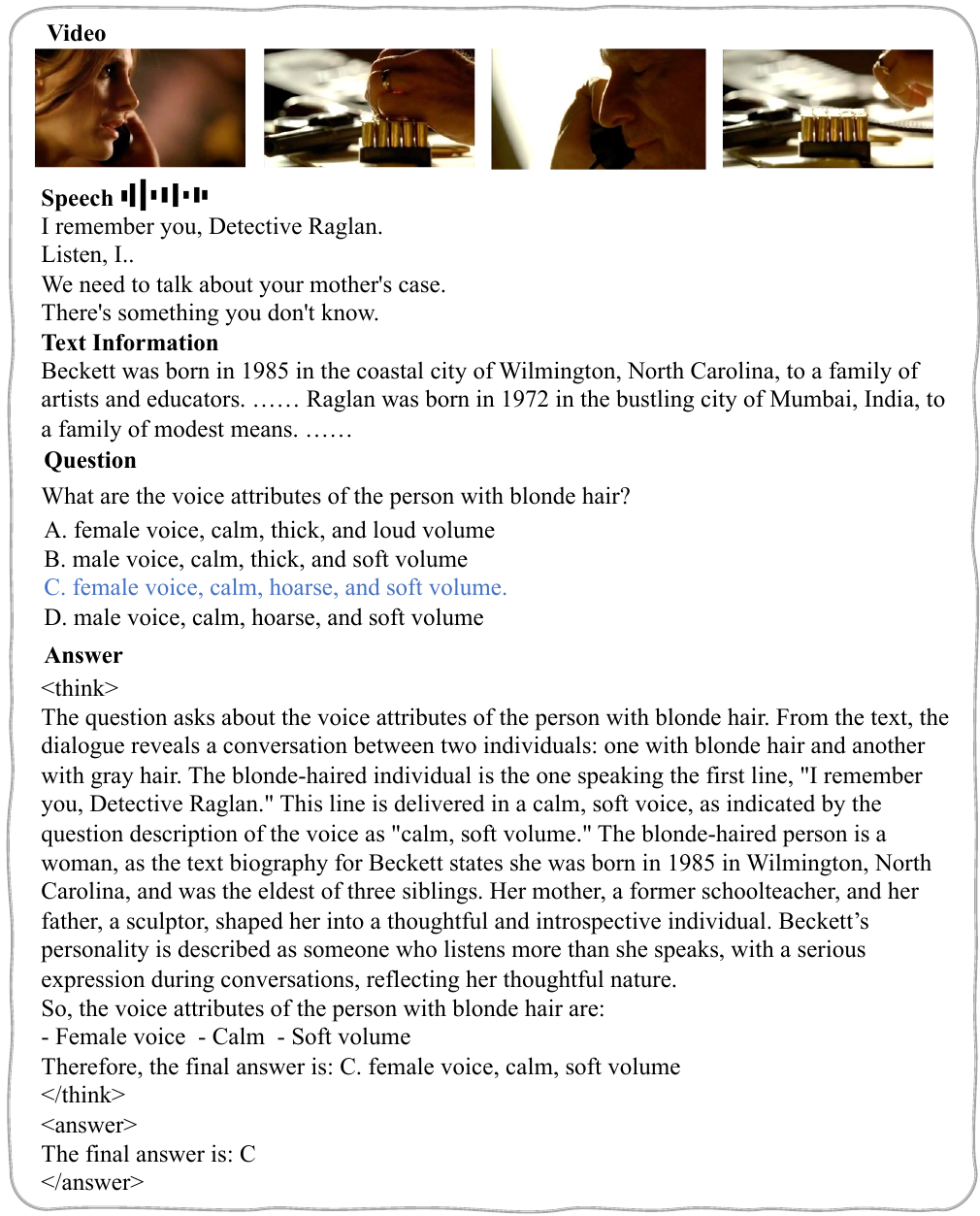}
    \caption{Case study on visual$\rightarrow$audio task via Qwen2.5-Omni-7b after training. The \textcolor{cot-blue}{blue} option denotes the right choice.}
    \label{fig:case1}
\end{figure*}

\begin{figure*}
    \centering
    \includegraphics[width=0.9\linewidth]{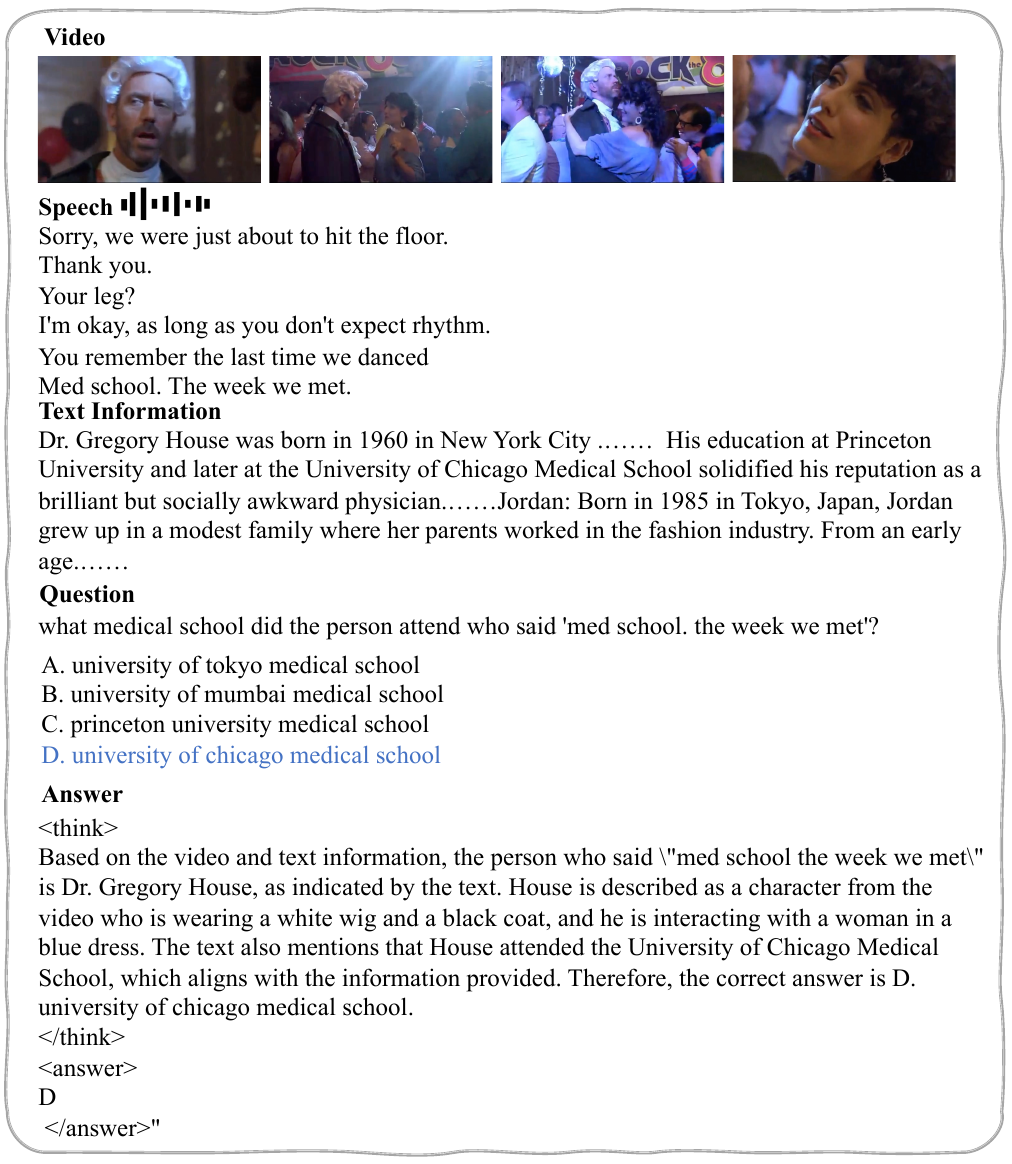}
    \caption{Case study on audio$\rightarrow$text task via Qwen2.5-Omni-3b after training. The \textcolor{cot-blue}{blue} option denotes the right choice.}
    \label{fig:case2}
\end{figure*}

To illustrate the effect of coreference-aware thinking patterns of training, we provide a CoT case in Figure~\ref{fig:cot_case} and two cases after training in Figure~\ref{fig:case1} and Figure~\ref{fig:case2}.

\section{Prompts Details}
For reproducibility, we show the prompts in Table~\ref{app-tab:modality_annotation}-Table~\ref{app-tab:grpo_prompt}, which consist of data annotation, in-context learning, and GRPO.

\begin{table*}[htbp]
\centering
\renewcommand{\baselinestretch}{0.9}
\begin{adjustbox}{max width=0.95\textwidth, center}
\begin{tabular}{l *{10}{S[table-format=2.2]}}
\toprule
\multicolumn{1}{c}{\multirow{2}{*}{\textbf{Model}}} &
\multicolumn{3}{c}{\textbf{Audio-centric}} &
\multicolumn{3}{c}{\textbf{Text-centric}} &
\multicolumn{3}{c}{\textbf{Visual-centric}} &
\multicolumn{1}{c}{\multirow{2}{*}{\textbf{Overall}}} \\
\cmidrule(lr){2-4} \cmidrule(lr){5-7} \cmidrule(lr){8-10}
& \textbf{Au} & \textbf{Au$\rightarrow$Te} & \textbf{Au$\rightarrow$Vi}
& \textbf{Te} & \textbf{Te$\rightarrow$Au} & \textbf{Te$\rightarrow$Vi}
& \textbf{Vi} & \textbf{Vi$\rightarrow$Au} & \textbf{Vi$\rightarrow$Te} & \\
\midrule
Baichuan-Omni        & 37.70 & 29.60 & 26.20 & 56.28 & 26.25 & 27.71 & 32.22 & 21.74 & 27.80 & 31.24 \\
Humanomni            & 33.23 & 41.09 & 46.96 & 54.88 & 31.56 & 45.45 & 56.07 & 31.30 & 39.00 & 41.53 \\
M2-Omni              & 44.73 & 37.36 & 32.91 & 73.02 & 28.12 & 46.32 & 53.14 & 31.30 & 54.05 & 43.19 \\
Ming-Omni            & 33.87 & 41.09 & 48.24 & 64.65 & 36.88 & 53.68 & 60.25 & 36.52 & 48.26 & 45.95 \\
MiniCPM              & 36.42 & 37.93 & 28.11 & 76.28 & 28.44 & 35.50 & 49.79 & 24.35 & 51.35 & 39.67 \\
Ola                  & 49.52 & 48.56 & 44.73 & 56.74 & 40.94 & 48.92 & 54.39 & 37.39 & 47.88 & 47.41 \\
Omni-r1              & 40.26 & 44.25 & 37.38 & 86.98 & 36.25 & 55.41 & 63.60 & 39.57 & 64.86 & 50.20 \\
OmniVinci            & 53.04 & 56.03 & 56.55 & 56.28 & 42.50 & 47.62 & 60.67 & 34.35 & 42.47 & 50.20 \\
Qwen2.5-Omni-3B      & 44.09 & 42.53 & 42.81 & 50.23 & 37.81 & 48.92 & 66.11 & 33.91 & 39.38 & 44.57 \\
Qwen2.5-Omni-7B      & 43.77 & 46.26 & 37.70 & 83.26 & 36.56 & 55.84 & 63.18 & 37.39 & 58.69 & 49.84 \\
Qwen3-Omni-30B-Cap   & 51.95 & 53.05 & 44.60 & 52.56 & 46.25 & 48.14 & 50.13 & 44.78 & 49.34 & 49.01 \\
Qwen3-Omni-30B-Ins   & 55.46 & 53.33 & 47.48 & 64.65 & 45.31 & 50.74 & 54.31 & 45.65 & 59.38 & 52.58 \\
Qwen3-Omni-30B-Thk   & 63.77 & 69.14 & 59.62 & 74.88 & 57.50 & 62.86 & 63.10 & 52.17 & 69.42 & 63.52 \\
\midrule
\multicolumn{11}{c}{\textbf{Average}} \\
\midrule
\multicolumn{1}{c}{--} & 45.22 & 46.17 & 42.56 & 65.44 & 38.03 & 48.24 & 55.92 & 36.19 & 50.15 & 46.83 \\
\bottomrule
\end{tabular}
\end{adjustbox}
\caption{Performance of different models across modalities via Chain-of-Thought. The Au, Te, and Vi are the abbreviations of Audio, Text, and Visual, respectively; X--Y denotes source-to-target modality (e.g., Au--Te: Audio-to-Text).}
\label{tab:cot_results}
\vspace{-0.8em}
\end{table*}

\begin{table*}[htbp]
\centering
\renewcommand{\baselinestretch}{0.9}
\begin{adjustbox}{max width=0.95\textwidth, center}
\begin{tabular}{l *{10}{S[table-format=2.2]}}
\toprule
\multicolumn{1}{c}{\multirow{2}{*}{\textbf{Model}}} &
\multicolumn{3}{c}{\textbf{Audio-centric}} &
\multicolumn{3}{c}{\textbf{Text-centric}} &
\multicolumn{3}{c}{\textbf{Visual-centric}} &
\multicolumn{1}{c}{\multirow{2}{*}{\textbf{Overall}}} \\
\cmidrule(lr){2-4} \cmidrule(lr){5-7} \cmidrule(lr){8-10}
& \textbf{Au} & \textbf{Au$\rightarrow$Te} & \textbf{Au$\rightarrow$Vi}
& \textbf{Te} & \textbf{Te$\rightarrow$Au} & \textbf{Te$\rightarrow$Vi}
& \textbf{Vi} & \textbf{Vi$\rightarrow$Au} & \textbf{Vi$\rightarrow$Te} & \\
\midrule
Baichuan-Omni        & 36.42 & 50.29 & 46.65 & 64.65 & 43.12 & 45.45 & 45.61 & 41.74 & 49.81 & 46.64 \\
Humanomni            & 42.17 & 44.25 & 58.47 & 60.47 & 38.44 & 46.75 & 61.92 & 41.30 & 49.03 & 48.62 \\
M2-Omni              & 42.49 & 48.28 & 53.04 & 72.56 & 45.94 & 48.92 & 52.72 & 47.39 & 53.67 & 50.93 \\
Ming-Omni            & 45.69 & 54.60 & 58.47 & 81.86 & 41.88 & 59.74 & 67.36 & 47.39 & 54.44 & 55.71 \\
MiniCPM              & 45.05 & 52.87 & 49.84 & 80.47 & 39.38 & 52.38 & 47.70 & 37.39 & 54.83 & 50.36 \\
Ola                  & 47.28 & 55.17 & 48.56 & 58.60 & 45.00 & 51.08 & 52.30 & 39.57 & 51.35 & 49.80 \\
Omni-r1              & 47.92 & 65.23 & 60.06 & 94.88 & 43.75 & 61.47 & 62.34 & 44.78 & 69.50 & 60.09 \\
OmniVinci            & 48.56 & 53.45 & 52.72 & 51.63 & 46.56 & 55.41 & 56.07 & 38.70 & 41.70 & 49.51 \\
Qwen2.5-Omni-3B      & 35.46 & 39.08 & 38.98 & 67.44 & 28.44 & 36.80 & 37.66 & 31.30 & 40.54 & 38.78 \\
Qwen2.5-Omni-7B      & 43.77 & 58.05 & 53.35 & 78.14 & 40.00 & 53.68 & 57.32 & 40.43 & 62.55 & 53.40 \\
Qwen3-Omni-30B-Cap   & 59.94 & 64.83 & 55.14 & 63.26 & 50.94 & 54.63 & 51.80 & 52.17 & 64.40 & 57.60 \\
Qwen3-Omni-30B-Ins   & 61.53 & 65.11 & 56.74 & 66.98 & 55.94 & 60.26 & 53.47 & 50.00 & 66.72 & 59.75 \\
Qwen3-Omni-30B-Thk   & 59.30 & 68.56 & 64.73 & 68.37 & 55.62 & 61.99 & 63.10 & 55.65 & 69.81 & 62.99 \\
\midrule
\multicolumn{11}{c}{\textbf{Average}} \\
\midrule
\multicolumn{1}{c}{--} & 47.35 & 55.37 & 53.60 & 69.95 & 44.23 & 52.97 & 54.57 & 43.68 & 56.03 & 52.63 \\
\bottomrule
\end{tabular}
\end{adjustbox}
\caption{Performance of different models across modalities via In-Context Learning. The Au, Te, and Vi are the abbreviations of Audio, Text, and Visual, respectively; X--Y denotes source-to-target modality (e.g., Au--Te: Audio-to-Text).}
\label{tab:icl_results}
\end{table*}

\label{sec:app-annotation-prompt}
\begin{table*}[htbp!]
\centering
\small
\setlength{\tabcolsep}{8pt}
\renewcommand{\arraystretch}{1.15}

\begin{tabularx}{\textwidth}{X}
\rowcolor[HTML]{343434}
\multicolumn{1}{c}{\color{white}\textbf{Prompt}}\\
\toprule
\rowcolor[HTML]{F0F0F0} \multicolumn{1}{c}{\textbf{Audio Annotation}}\\
Is the voice smooth? \\
Is the voice rough?\\
Is the voice nasal?\\
Is the pitch high-pitched?\\
Is the pitch low-pitched?\\
Does the speech sound happy?\\
Does the speech sound angry?\\
Does the speech sound calm?\\
\rowcolor[HTML]{F0F0F0} \multicolumn{1}{c}{\textbf{Single Video Frame Annotation}}\\
Describe the sequence of images in a detailed and continuous narrative, focusing on the following key aspects:\\
1. People:
    a.Describe the individuals in each image (clothing, appearance, expressions, body language).
    b.Track any changes in their appearance or actions across images.\\
2.Background and Environment:
    a.Include detailed descriptions of the setting (e.g., buildings, nature, objects).
    b.Provide spatial context (left, right, foreground, background).
    c.Mention any visible objects and their interaction with people.\\
3.Sequence and Time:
    a.Follow a chronological order, indicating how the scene evolves.
    b.Ensure the narrative flows naturally, showing changes over time.
    c.Do not descibe this image by image.\\
Make sure the description is rich in detail, including both human and environmental elements, while maintaining clarity and coherence \\
\rowcolor[HTML]{F0F0F0} \multicolumn{1}{c}{\textbf{Multiple Video Frames Combination}}\\
You will receive an overall description of a video along with descriptions of individual segments of the video. Your task is to integrate these into a comprehensive and detailed description. Follow the rules below:\\
        1. **Overall Video Description**: This description provides the general background, theme, and main content of the video.\\
        2. **Segment Descriptions**: These provide detailed descriptions of each segment of the video, which may contain specific events or scenes.\\
        Please generate the final, complete description as follows:\\
        - Integrate the overall description with the individual segment descriptions, ensuring the final output covers all aspects of the video.\\
        - If there are any inconsistencies between the overall description and the segment descriptions, ignore the inconsistent parts and only retain the consistent and accurate information.\\
        - The final description should be coherent, and events should be presented in chronological order, ensuring a logical flow throughout.\\
\rowcolor[HTML]{F0F0F0} \multicolumn{1}{c}{\textbf{Person Description}}\\
There is a video with one person is speaking. Please describe the \{\textcolor{blue}{external knowledge}\} in detail who is speaking, including characteristics and clothing. \\
1.Do not describe the background. \\
2.Only output the \{\textcolor{blue}{external knowledge}\}. \\
3.Use '\{\textcolor{blue}{external knowledge}\}' as the beginning.\\
4.If there is no person speaking, directly output <failure>. \\
\rowcolor[HTML]{F0F0F0} \multicolumn{1}{c}{\textbf{Text Biography Generation}}\\
Please generate detailed biographies for the following characters based on their provided background information (name + apparel). For each character, create a unique and detailed life story, highlighting their experiences, relationships, challenges, and achievements, including but not limited to: \\
        1. **Place of birth and upbringing**: Where they were born and raised, family environment, education background, etc.\\
        2. **Significant life events**: Major life events or turning points that shaped their life.\\
        3. **Career and achievements**: Their career path, important achievements, and contributions.\\
        4. **Relationships**: Key relationships with others, such as family, friends, enemies, or partners.\\
        5. **Personality and psychological development**: Their personality traits and any psychological or emotional growth.\\
        Ensure the biography covers all provided background details, and the name used in the biography must match the one provided.\\
        The format is: Name + Biography.\\
        <background information>:\\
\bottomrule
\end{tabularx}
\caption{Prompt for modality annotation.}
\label{app-tab:modality_annotation}
\end{table*}
\begin{table*}[htbp!]
\centering
\small
\setlength{\tabcolsep}{8pt}
\renewcommand{\arraystretch}{1.15}

\begin{tabularx}{\textwidth}{X}
\rowcolor[HTML]{343434}
\multicolumn{1}{c}{\color{white}\textbf{Prompt}}\\
\toprule
\rowcolor[HTML]{F0F0F0} \multicolumn{1}{c}{\textbf{Audio Double Check}}\\
Analyze the descriptions and identify any audio that stands out due to unique features such as:\\
        1.Gender: Unusually male or female voice.\\
        2.Emotion: Distinct emotional tone (e.g., anger, sadness).\\
        3.Audio Quality: Notable differences in speed, pitch, or volume.\\
        4.Other Unusual Traits: Background noise, interruptions, etc.\\
        If there are any audio is different, output this name(s) with the distinct feature and briefly explain what makes it different.\\
        The format is: <name + order>  + <distinct feature> + <reason>.\\
        If all the descriptions are similarly, output <failure>.\\
\rowcolor[HTML]{F0F0F0} \multicolumn{1}{c}{\textbf{Video Double Check: Video Description}}\\    
Describe the video in a detailed and continuous narrative, focusing on the following key aspects:\\
1. People:
    a.Describe the individuals in each image (clothing, appearance, expressions, body language).
    b.Track any changes in their appearance or actions across images.\\
2.Background and Environment:
    a.Include detailed descriptions of the setting (e.g., buildings, nature, objects).
    b.Provide spatial context (left, right, foreground, background).
    c.Mention any visible objects and their interaction with people.\\
Make sure the description is rich in detail, including both human and environmental elements, while maintaining clarity and coherence \\
\rowcolor[HTML]{F0F0F0} \multicolumn{1}{c}{\textbf{Video Double Check: Double Check}}\\    
You are a strict Video Description Consistency Checker. You will be given two texts, description1 and description2, which are upposed to describe the same video. Your job is to determine whether they are consistent at the core factual level. If there is any fundamental mismatch, output exactly failure. Otherwise output exactly pass.\\
Definition of Fundamental Conflict (any one $\Rightarrow$ failure):\\
1) People conflicts: significantly different number of main people; clearly contradictory identity/type (e.g., adult man vs.\ child girl); mutually exclusive key actions/events (e.g., driving vs.\ cooking at a desk).\\
2) Background conflicts: clearly different main scene type (e.g., kitchen vs.\ highway); strongly contradictory environment (e.g., heavy rain outdoors vs.\ quiet indoor office); incompatible main objects/interactions.\\
3) Narrative conflicts: overall story/sequence cannot reasonably align as the same video; descriptions refer to different scenarios/storylines.\\

Allowed Differences (do NOT count as fundamental conflicts):\\
1) Different level of detail (one richer, one concise).\\
2) Minor omissions or reordering without contradicting main events.\\
3) Small secondary-detail differences (e.g., unmentioned background object, ambiguous colors/positions) when core people/setting/actions match. \\

\rowcolor[HTML]{F0F0F0} \multicolumn{1}{c}{\textbf{Person Double Check}}\\    
Analyzes the consistency of descriptions for each name and outputs whether they are correct or not.\\
            Rules:\\
            1. If the same name has multiple completely different descriptions, choose the description that appears most frequently as the final description.\\
            2. If any description contains terms like 'failed', 'cannot describe', or anything that indicates the description is not valid, directly output <Failed>.\\
            3. If different names have the same description (roughly similar),  directly output <Failed>.\\
            4. If there no failed casess, output the name and its description. The format is: name + description.\\
\bottomrule
\end{tabularx}
\caption{Prompt for modality annotation double check.}
\label{app-tab:modality_annotation_db}
\end{table*}
\begin{table*}[htbp!]
\centering
\small
\setlength{\tabcolsep}{8pt}
\renewcommand{\arraystretch}{1.1}

\begin{tabularx}{\textwidth}{X}
\rowcolor[HTML]{343434}
\multicolumn{1}{c}{\color{white}\textbf{Prompt}}\\
\toprule
\rowcolor[HTML]{F0F0F0} \multicolumn{1}{c}{\textbf{QA Generation}}\\
Design a question where the model infers **factual description** from **visual actions**, using the following information:\\
- **Overall Descriptions**\\
- **person biography**\\
- **object biography**\\
Consider these aspects:\\
1. **Visual Actions**:\\
   - These include body language, gestures, and facial expressions.\\
2. **Factual description**: \\
   - These refer to stated details such as names, dates, locations, occupations, and major life events (e.g., education, career milestones, relationships).\\
Design a question to infer **factual description** from **visual actions**.\\
If you cannot design an appropriate question, directly output ``<failure>''.\\
\rowcolor[HTML]{F0F0F0} \multicolumn{1}{c}{\textbf{QA Double Check: Unique}}\\
Task: Read the Question, the text biography. The Question ask somethings about a person or a object. Your job is to find the exact name(s) of the character(s) in the visual description who satisfy the description.\\
Input:\\
Question: \{\textcolor{blue}{question}\}\\
Visual description: \{\textcolor{blue}{video des}\}\\
Person description: \{\textcolor{blue}{person des}\}\\
Output whether only one the person or object who match the description in the Question.\\
Output: yes or no \\
\rowcolor[HTML]{F0F0F0} \multicolumn{1}{c}{\textbf{QA Double Check: Answerable}}\\
                Task: Read the Question, the Answer, the text biography, the Person description, the Visual description. Your job is to determine whether this Answer is correct based only on the provided Context.\\
                Input:\\
                Question: \{\textcolor{blue}{question}\}\\
                Answer: \{\textcolor{blue}{answer}\}\\
                text biography: \{\textcolor{blue}{person biography}\}\\
                Visual description: \{\textcolor{blue}{video des}\}\\
                Person description: \{\textcolor{blue}{person des}\}\\

                Rules: \\
                (1) You should first indentify the person or object according to the question, visual description.\\
                (2) Secondly, you should extract the name of the person according to the person description or object according to the visual description.\\
                (3) Thirdly, you should find this person in text biography or object in object biography.\\
                (4) Then, you should answer this question based on the biography.\\ 
                (5) Finally, you should compare the similarity between two answers and determine whether the answer can be correct to this quesiton.\\ 
                
                Output:\\
                correct or non-correct.\\
\rowcolor[HTML]{F0F0F0} \multicolumn{1}{c}{\textbf{Option Generation}}\\
Task: Read the Question, the Answer, the text biography, the Person description, the Visual description. Your job is to determine whether this Answer is correct based only on the provided Context.\\
                Input:\\
                Question: \{\textcolor{blue}{question}\}\\
                Answer: \{\textcolor{blue}{answer}\}\\
                text biography: \{\textcolor{blue}{person biography}\}\\
                Visual description: \{\textcolor{blue}{video des}\}\\
                Person description: \{\textcolor{blue}{person des}\}\\
                Rules: \\ 
                (1) You should first indentify the person or object according to the question, visual description.\\ 
                (2) Secondly, you should extract the name of the person according to the person description or object according to the visual description.\\ 
                (3) Thirdly, you should find this person in text biography or object in object biography.\\ 
                (4) Then, you generate other three <incorrect answer>, the answer can be obtained from the biography or not from.\\ 
                (5) Finally, you should ensure the <incorrect answer> are wrong for this question.\\ 
                
                Output:\\ 
                <incorrect answer>: xxx \\ <incorrect answer> xxx \\ <incorrect answer> xxx \\
\bottomrule
\end{tabularx}
\caption{Prompt for QA generation. We use the visual$\rightarrow$text questions in visual actions as a case study.}
\label{app-tab:qa_generation}
\end{table*}

\begin{table*}[htbp!]
\centering
\small
\setlength{\tabcolsep}{8pt}
\renewcommand{\arraystretch}{1.1}
\begin{tabularx}{\textwidth}{X}
\rowcolor[HTML]{343434}
\multicolumn{1}{c}{\color{white}\textbf{Prompt}}\\
\toprule
\rowcolor[HTML]{F0F0F0} \multicolumn{1}{c}{\textbf{QA Generation}}\\
            Task:\\
            You will produce detailed step-by-step reasoning (``Reasoning Details'') that explains and leads to the correct Answer.\\
            Very important constraints:\\
            - You must pretend that ALL useful information comes ONLY from:\\
            (a) what is seen in the video (visual content), and\\
            (b) what is known from the biographies (person biography and object biography).\\
            - You may explicitly refer to the Question and to the person/object biographies.\\
            - You may internally use the Visual description and Person description to help you reconstruct what could be seen in the video\\
            and how it connects to the biographies, but you MUST NOT mention or imply that you used any descriptions, text, metadata,\\
            or annotations as special inputs.\\
            - You MUST NOT say that the Answer was given to you in advance.\\
            - You secretly know the correct Answer from the input, and your reasoning must naturally lead to exactly this Answer,\\
            but it should look like you inferred it only from watching the video and reading the biographies.\\
            Inputs:\\
            - Question: \{\textcolor{blue}{question}\}\\
            - Answer: \{\textcolor{blue}{answer}\}\\
            - text biography: \{\textcolor{blue}{ biography}\}\\
            - Visual description: \{\textcolor{blue}{video des}\}\\
            - Person description: \{\textcolor{blue}{person des}\}\\
            Interpretation Rules:\\
            1) You MUST act as if all visual information (who/what appears, appearance, position, actions, environment, etc.)
            comes directly from watching the video.
            2) You MUST act as if all background/identity information about people comes from the person biography.\\
            3) You MUST act as if all background/identity information about objects comes from the object biography.\\
            4) The Visual description and Person description are only internal tools to help you reconstruct
            what WOULD HAVE BEEN seen in the video and how it connects to the biographies. Do not mention or allude to them.\\
            5) In your reasoning, always frame information as observations from:\\
            - watching the video, and\\
            - reading the relevant biography (person or object).\\
            Reasoning Details Rules:\\
            1) First, determine whether the Question is about a person or an object, and identify which specific person or object it refers to,
            as if you are using only the video content (internally you may rely on the Visual description, but never mention it).\\
            2) Second, if the Question is about a person, infer their name and identity as if you know it from the person’s biography
            (internally using the Person description + person biography); if it is about an object, infer its identity/type as if you
            recognize it from the video and the object biography.\\
            3) Third, match this person to the person biography, or this object to the object biography, and explain how their
            background or properties are relevant to the Question.\\
            4) Then, using the relevant biography information (person or object) together with what is seen in the video,
            logically derive and justify the correct Answer step by step. Make the reasoning explicit and multi-step, not just one short sentence.\\
            5) Your final conclusion MUST exactly match the hidden Answer above, but you MUST NOT reveal that this Answer was given to you.\\
            6) The Reasoning Details must be entirely in English.\\
            Output format:\\
            <Reasoning Details>:\\
            Your step-by-step reasoning here\\
            <Final Answer>:\\
            State the final answer here, matching the hidden Answer \\
\bottomrule
\end{tabularx}
\caption{Prompt for CoT generation. We use the visual$\rightarrow$text questions as a case study.}
\label{app-tab:cot_generation}
\end{table*}
\label{sec:app-icl-prompt}
\begin{table*}[htbp!]
\centering
\small
\setlength{\tabcolsep}{8pt}
\renewcommand{\arraystretch}{1.1}
\begin{tabularx}{\textwidth}{X}
\rowcolor[HTML]{343434}
\multicolumn{1}{c}{\color{white}\textbf{Prompt}}\\
\toprule
\rowcolor[HTML]{F0F0F0} \multicolumn{1}{c}{\textbf{In-Context Learning}}\\
Answer the question based on the video and the text information in options A/B/C/D.\\

You are REQUIRED to first think step by step and write out your reasoning, and THEN give the final answer.\\

You MUST strictly follow this output format:\\

<think>\\
Your detailed step-by-step reasoning in natural language. Explain how you identify the person or object from the question and video, describe its visual appearance, find the Additional Information using that description, and then use the Additional Information to choose the correct option.\\
</think>\\
Final answer only: A / B / C / D\\

Here are the rules you MUST follow:\\
(1) Carefully read the question and observe the video. First, use them to identify the person or object mentioned in the question.\\
(2) Secondly, describe the visual appearance of this person or object in the video.\\
(3) Thirdly, find the Additional Information about this person or object according to the visual description.\\
(4) Then, answer the question based on the Additional Information.\\
(5) All identification, description, retrieval of Additional Information, and reasoning steps MUST be written ONLY inside the <think> section.\\
(6) Outside the <think> tags, you MUST output ONLY the final answer (for example, just “A”, “B”, “C”, or “D”) with no extra words, no explanation, and no additional text.\\
(7) Do not output anything before <think> or after the final answer. Follow exactly this structure:\\
<think> ... </think>\\
A/B/C/D\\

If you do NOT follow this format, your answer will be considered incorrect.\\
\bottomrule
\end{tabularx}
\caption{Prompt for in-context learning. We use the visual$\rightarrow$text questions as a case study.}
\label{app-tab:icl_prompt}
\end{table*}
\label{sec:app-GRPO-prompt}
\begin{table*}[htbp!]
\centering
\small
\setlength{\tabcolsep}{8pt}
\renewcommand{\arraystretch}{1.1}
\begin{tabularx}{\textwidth}{X}
\rowcolor[HTML]{343434}
\multicolumn{1}{c}{\color{white}\textbf{Prompt}}\\
\toprule
\rowcolor[HTML]{F0F0F0} \multicolumn{1}{c}{\textbf{Contextual Consistency}}\\
You are assessing how well the 'hypothesis' text covers the key information from the 'reference' text. Differences in wording or extra details in the 'hypothesis' are fine if the 'reference's' main points are included.:\\

Score based on this coverage:\\

5 points : Hypothesis clearly and accurately reflects significant core themes or key aspects of the reference. It demonstrates a good understanding of a substantial part of the reference material.\\
4 points : Hypothesis reflects some important themes or aspects of the reference. The connection is evident, though perhaps not as comprehensive or central as a 5.\\
2 points : Hypothesis shows a recognizable connection to themes or aspects of the reference, but it might be more superficial, focus on less central points, or only partially grasp a key aspect.\\
1 points : Hypothesis has a tenuous or very limited connection to the reference. It might touch on a peripheral detail or a heavily reinterpreted aspect, but largely misses the main substance.\\
0 points : Hypothesis does not reflect any significant themes or key aspects of the reference, or is on a completely different topic.\\

Example analysis process:\\

Identify main themes and key aspects in 'reference'.
Determine if 'hypothesis' connects to or discusses any of these themes/aspects from 'reference'.\\
Judge the strength and relevance of this connection. Is a core part of the 'reference' reflected?\\
Differences are expected; evaluate if the 'hypothesis' still meaningfully reflects some key part of the 'reference'.\\
Assign score based on how well a significant aspect is reflected.\\

reference: \{\textcolor{blue}{reference}\}\\
hypothesis: \{\textcolor{blue}{hypothesis}\}\\

only return the score number:\\

\rowcolor[HTML]{F0F0F0} \multicolumn{1}{c}{\textbf{Logical Coherence}}\\
            Please evaluate whether the reasoning path performs effective cross-modal reasoning from visual video content to text (Additional Information), and give a score from 0 to 3.\\

            Scoring criteria (1 point for each satisfied item):\\

            1. Question-guided visual localization (1 point):
            The reasoning interprets the question and uses it to identify and localize the relevant person/object in the video, including describing its visual appearance, rather than directly searching the text.\\

            2. Text retrieval via visual description (1 point):
            Using this visual description, the reasoning explicitly finds the corresponding Additional Information about this person/object in the text (e.g., matching name or described appearance), clearly linking video appearance to a specific text entry.\\

            3. Answer grounded in retrieved text (1 point):
            The reasoning then uses the retrieved textual Additional Information as the key evidence to select the final option A/B/C/D, with the answer clearly supported by that text and without major unsupported assumptions or contradictions.\\

            Scoring rule:\\
            - Award 1 point for each criterion that is clearly satisfied.\\
            - The final score is an integer from 0 to 3.\\

            reasoning path: \{\textcolor{blue}{hypothesis}\}\\

            Only return the score number (an integer from 0 to 3):\\
\bottomrule
\end{tabularx}
\caption{Prompt for GRPO. We use the visual$\rightarrow$text questions as a case study.}
\label{app-tab:grpo_prompt}
\end{table*}

\end{document}